\documentclass[10pt,twocolumn,letterpaper]{article}

\usepackage{cvpr}              
\definecolor{cvprblue}{rgb}{0.21,0.49,0.74}
\usepackage[pagebackref,breaklinks,colorlinks,allcolors=cvprblue]{hyperref}
\usepackage{adjustbox}
\usepackage{multirow}
\usepackage{multicol}
\usepackage{graphicx}
\usepackage{caption}
\usepackage{amsmath}
\usepackage{amssymb}
\usepackage{algorithm}
\usepackage{makecell}
\usepackage{algpseudocode}
\usepackage{amsfonts}
\def\paperID{4658} 
\def\confName{CVPR}
\def\confYear{2025}
\newcommand{\minisection}[1]{\vspace{0.005in} \noindent {\bf #1}}

\usepackage{colortbl}
\definecolor{LightPurple}{rgb}{0.88,0.88,1}

\title{Language Guided Concept Bottleneck Models for \\ Interpretable Continual Learning}

\author{
 Lu Yu \textsuperscript{1}
  \hspace{15pt} Haoyu Han \textsuperscript{1} \hspace{15pt} Zhe Tao \textsuperscript{1} \hspace{15pt} Hantao Yao \textsuperscript{2}
  \hspace{15pt} Changsheng Xu \textsuperscript{3}
     \\\\
     \hspace{-10pt} 
     \textsuperscript{1}School of Computer Science and Engineering, Tianjin University of Technology  \\
     \hspace{-10pt} 
     \textsuperscript{2}School of Information Science and Technology, University of Science and Technology of China \\
     \hspace{-10pt} 
     \textsuperscript{3}State Key Laboratory of Multimodal Artificial Intelligence Systems, \\ Institute of Automation, University of Chinese Academy of Sciences
     \\
     \texttt{\{luyu@email, fishercat@stud, tz@stud\}.tjut.edu.cn}, \\ \texttt{yaohantao@ustc.edu.cn, csxu@nlpr.ia.ac.cn}
}

\begin{document}
\maketitle
\captionsetup[figure]{skip=5pt}
\begin{abstract}

Continual learning (CL) aims to enable learning systems to acquire new knowledge constantly without forgetting previously learned information. CL faces the challenge of mitigating catastrophic forgetting while maintaining interpretability across tasks. Most existing CL methods focus primarily on preserving learned knowledge to improve model performance. However, as new information is introduced, the interpretability of the learning process becomes crucial for understanding the evolving decision-making process, yet it is rarely explored. In this paper, we introduce a novel framework that integrates language-guided Concept Bottleneck Models (CBMs) to address both challenges. Our approach leverages the Concept Bottleneck Layer, aligning semantic consistency with CLIP models to learn human-understandable concepts that can generalize across tasks. By focusing on interpretable concepts, our method not only enhances the model’s ability to retain knowledge over time but also provides transparent decision-making insights. We demonstrate the effectiveness of our approach by achieving superior performance on several datasets, outperforming state-of-the-art methods with an improvement of up to 3.06\% in final average accuracy on ImageNet-subset. Additionally, we offer concept visualizations for model predictions, further advancing the understanding of interpretable continual learning. Code is available at \url{https://github.com/FisherCats/CLG-CBM}.

\end{abstract}

\section{Introduction}
In dynamic and ever-changing environments, models trained on fixed datasets often struggle to adapt to new tasks or data without losing previously acquired knowledge, facing the challenge known as catastrophic forgetting~\cite{mccloskey1989catastrophic}. Continual learning~\cite{de2021continual,masana2022class,zhou2024class} addresses this limitation by allowing systems to retain and incrementally build upon past knowledge, resulting in greater flexibility. This capability is especially critical in real-world applications like robotics, autonomous systems, and personalized healthcare, where models must continuously learn from ongoing experiences and data streams to stay effective. Additionally, continual learning improves the transfer of knowledge across tasks, minimizing the need for frequent retraining and enhancing computational efficiency. These benefits make continual learning a promising approach for advancing intelligent and adaptive AI systems in real-world.

Existing continual learning methods~\cite{lwf,ER,RPSnet,SDC,l2p} focus on mitigating the issue of catastrophic forgetting yet the underlying mechanisms behind this phenomenon remain largely unexplained. As these models continuously update their knowledge, it becomes crucial to understand what they learn and how they retain previous information to prevent unintended behavior and enhance human interpretability. Furthermore, understanding the internal mechanisms of continual learning models can guide improvements in their design, leading to more robust and effective systems.

Despite the importance of this challenge, few papers have explored the interpretability of continual learning models. ICICLE~\cite{ICICLE} is one such work, which interprets concept drift in continual learning through a prototypical-part network and introduces interpretability regularization to minimize changes in prototype similarities.

Concept Bottleneck Models (CBMs)~\cite{CBM, wang2023learning} are a form of interpretable machine learning model that enhance transparency by organizing the learning process around human-understandable concepts. Rather than directly mapping inputs to predictions, CBMs incorporate an intermediate `bottleneck' layer composed of high-level, explicitly defined concepts that are meaningful to humans. Some recent research~\cite{LABO,LM4CV,LFCBM,shang2024incremental} integrates language-guided approaches with CBMs to further align the model’s decision-making process with human cognition, particularly for tasks involving abstract or semantic knowledge. Leveraging the inherent interpretability and structured knowledge representation of CBMs, this paper investigates their potential application to continual learning, with the goal of enhancing both interpretability and adaptability while ensuring the retention of previously acquired knowledge.

With the rapid advancements in large-scale multimodal models, we can harness their robust zero-shot encoding capabilities to establish a strong foundation for extracting high-quality features. In this paper, we integrate the pre-trained CLIP~\cite{CLIP} model to obtain bottleneck concepts within a continual learning framework. Specifically, we utilize pre-trained language models (e.g., ChatGPT~\cite{brown2020language}) to generate human-understandable concepts for each category. These concepts are subsequently encoded into concept embeddings through the CLIP text encoder, after which the most informative and expressive concepts are selected to form a task-specific bottleneck. We construct the Concept Bottleneck Layer (CBL) by aligning the concept score matrix with CLIP concept activiation matrix, enhancing model's interpretability. Furthermore, we leverage semantic knowledge to augment prototypes, effectively mitigate catastrophic forgetting. Finally, our approach achieves superior performance on seven benchmark datasets, consistently maintaining interpretability throughout the  process. 

Our main contributions can be summarized as follows:
\begin{itemize}
    \item 
    We introduce a novel framework that leverages language-guided Concept Bottleneck Models to enhance both interpretability and the ability to mitigate catastrophic forgetting in continual learning.
    \item 
    We construct the Concept Bottleneck Layer by aligning semantic consistency with CLIP models, enabling the learning of human-understandable concepts across tasks.
    \item 
    Our method outperforms state-of-the-art methods on the several datasets. Additionally, we provide concept visualizations for model predictions to enhance the understanding of interpretable continual learning.
\end{itemize}

\section{Related Work}   

\subsection{Continual Learning}
There are three main types of existing continual learning methods. \textit{Regularization-based} methods~\cite{lwf,LWM,aljundi2018memory,EWC,zenke2017continual,PASS,Bilateral_MAE,TASS,RAPF,CIL4LWNetworks,zhai2024fine,Gomez-Villa_2022_CVPR} typically alleviate catastrophic forgetting either by constraining the model's outputs through knowledge distillation or by penalizing changes to critical parameters based on their assessed importance. \textit{Rehearsal-based} methods~\cite{icarl,hou2019learning,ER, chaudhryefficient, CLFD,yang2023continual} involve replaying data from previous tasks by storing a subset of past data or by generating synthetic data, to prevent forgetting. The replayed data can be used at training stage with the current task data, maintaining the model's performance on previous tasks. \textit{Architecture-based} methods~\cite{DER,PACKnet,yoon2018lifelong} adapt the model's structure dynamically as new tasks are introduced, such as activating different part of model parameters for different tasks, using modular designs to compartmentalize learning and minimize interference between tasks.

Methods based on pre-trained models (PTMs) have made significant progress recently by leveraging robust features extracted from large-scale PTMs. Among these methods, approaches~\cite{l2p,dualprompt,codaprompt,CPP,LAE} based on \textit{Parameter-Efficient Tuning} typically construct and train a set of PEF modules, improving performance while reducing computational costs. \textit{Representation-based} methods~\cite{con-clip,zhang2023slca,EASE}, on the other hand, usually utilize the robust features of PTMs to construct classifiers.

The methods discussed above effectively alleviate the problem of catastrophic forgetting, a central challenge in continual learning. However, the decision-making process of those methods is still a black box and not transparent, and the rationality for model predictions shifts over time due to catastrophic forgetting, making it difficult to improve interpretability in continual learning. To address this issue, ICICLE~\cite{ICICLE} has explored interpretable continual learning methods by introducing a prototypical parts-based approach, they further proposed interpretability regularization and proximity-based prototype initialization, regularizing model to activate similar prototypical parts learned previously to mitigate Interpretability Concept Drift (ICD). Although ICICLE provides interpretability, it substantially limits the plasticity of the model, highlighting the need for methods that are both interpretable and effective. 

\subsection{Interpretable Deep Learning}      
There are two types of interpretable models that have been well-developed in the domain of deep learning explanations: post-hoc models and self-explainable models. Post-hoc models aim to elucidate the reasoning process of black-box methods, using techniques such as saliency maps~\cite{Gradcam, selvaraju2019taking}, concept activation vectors~\cite{kim2018interpretability,yeh2020completeness,CBM}, counterfactual examples~\cite{abbasnejad2020counterfactual,goyal2019counterfactual,mothilal2020explaining} and prototype similarity~\cite{DPPnet,PPnet,rymarczyk2021protopshare}, etc. While self-explainable models~\cite{alvarez2018towards,nauta2021neural,kim2022vit} aim to make the model intrinsically interpretable by faithfully representing the entire classification behavior, they are often more complex than post-hoc models.

Among these interpretable models, Concept Bottleneck Models~\cite{CBM,PCBM,LABO,LM4CV,CoopCBM} provide explanations of the model's decision-making process in a straightforward manner. CBMs are designed to be interpretable, incorporating an intermediate Concept Bottleneck Layer (CBL), where each neuron represents a high-level, expressive concepts. Image features are projected into Concept Space to acquire concept score vectors, which are utilized to make final classification. Recent research~\cite{LABO,LM4CV,LFCBM} has integrated textual knowledge with CBMs, addressing the challenges of obtaining class concepts and annotations by querying LLMs (e.g. chatGPT) and utilizing VLM (e.g. CLIP) to encode images and concepts. The similarity between the image and the concepts is calculated to determine the probability of each concept's presence in the image, producing a similarity vector that serves for interpretation and classification, making the decision making process more transparent.

\section{Preliminary}
\subsection{Class-Incremental Learning}
We focus on class incremental learning (CIL) in this paper, where the learning process unfolds across multiple tasks. For CIL scenario with $n$ tasks, let $\mathcal{X}_t$ denote the input space, and $\mathcal{Y}_t$ represent the set of labels or classes observed by the model at task $t$. The data associated with task $t$ can be denoted as $\mathcal{D}_t = \{(x_i, y_i)\}_{i=1}^{N_t}$, where $x_i \in \mathcal{X}_t$ and $y_i \in \mathcal{Y}_t$. At each task $t$, the model is introduced to a new set of classes $\mathcal{Y}_t$, which it learns to recognize these new classes while retaining knowledge of previously encountered classes. 

The sequence of tasks is denoted as $\mathcal{T} = \{1, 2, \dots, n\}$. In CIL, classes included in separate tasks are non-overlapping, where $\mathcal{Y}_{i} \cap \mathcal{Y}_{j} = \emptyset \quad (\forall \ i \neq j)$. The entire class space after task $t$ is represented as: $\mathcal{Y}_{1:t} = \mathcal{Y}_1 \cup \mathcal{Y}_2 \cup \dots \cup \mathcal{Y}_t.$ During training for task $t$, the model has access only to the current task's dataset $\mathcal{D}_t$, but it needs to predict labels from the cumulative set of classes $\mathcal{Y}_{1:t}$.

\subsection{Concept Bottleneck Models}
Compared to conventional deep learning architectures, Concept Bottleneck Models (CBMs)~\cite{CBM, wang2023learning} introduce a Concept Bottleneck Layer (CBL) positioned between feature extractor and classifier. Concept Bottleneck Layer comprises neurons that represent human-understandable concepts. These neurons convert image features into concept scores, which quantify the degree of alignment between the image and predefined concepts. The classifier then leverages these concept scores to make predictions, thereby enhancing the interpretability of the decision-making process. Language-Guided CBMs~\cite{LABO,LM4CV,LFCBM,shang2024incremental} extend this framework by integrating natural language processing (NLP) with concept-based modeling to further enhance interpretability and performance. A notable example is the CLIP-based CBMs \cite{LABO,LM4CV}, which employs the pre-trained CLIP model as its backbone. For instance, CLIP-based CBM method \cite{LM4CV} initially gathers relevant concepts by querying ChatGPT, followed by a concept selection module which is developed to identify the most informative and discriminative concepts. Specifically, a simple Multi-Layer Perceptron is trained in \cite{LM4CV} to capture the semantic knowledge embedded in images, guided by both Cross-Entropy loss and Mahalanobis loss. The acquired semantic knowledge is subsequently utilized for concept selection, constructing a concept pool $\mathcal{C}$ to facilitate model interpretability. The text features of $\mathcal{C}$ serves as the CBL in CLIP-based CBMs, and the size of CBL is determined by the quantity of selected concepts for all seen tasks. 

The CLIP concept activation matrix is computed as the dot product between the image features $f_I(\mathcal{X})$ and text features $f_T(\mathcal{C})$ as follows, enabling the model to align visual inputs with their corresponding semantic concepts:
\begin{equation}
    E_{clip} = f_I(\mathcal{X}) \cdot f_T(\mathcal{C})^\top\label{eq:clip_activation_matrix}
\end{equation}
The final predication $\hat{y}$ can be calculated as follows: 
\begin{equation}
    \hat{y} = \operatorname{argmax}\ \sigma(E_{clip} \cdot W_l^\top),\label{eq:pred}
\end{equation}
where $W_l$ refers to the weight of classifier, $\sigma$ refers to sigmoid function.

\section{Method}
\subsection{Overview Framework}
\begin{figure*}[t]
\centering
\includegraphics[width=0.8\textwidth]{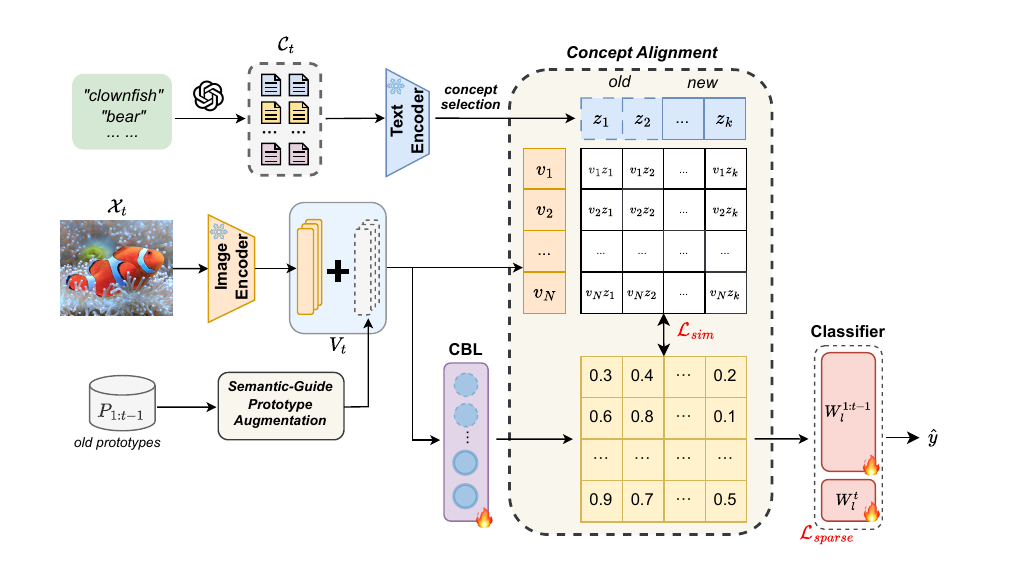}
\caption{
The framework of our method for task \textit{t}. The concept alignment module aligns the concept score matrix with the CLIP concept activation matrix. The semantic-guided prototype augmentation module leverages semantic knowledge to identify the most semantically similar class within $\mathcal{Y}_t$, generating pseudo-features for previously learned classes to mitigate catastrophic forgetting.
}
\label{fig:framework}
\vspace{-2mm}
\end{figure*}

We introduce Language-Guided Concept Bottleneck Models as an efficient framework for achieving interpretable continual learning. CBMs are designed to learn human-understandable intermediate representations, known as concepts, which serve as the basis for predictions and enhance transparency. However, while CBMs improve interpretability, applying them directly in a continual learning setting can still lead to catastrophic forgetting. To mitigate this, we propose a \textit{semantic-guided prototype augmentation module} to generate pseudo-features for old classes by leveraging semantic similarity with new data. Additionally, we introduce a \textit{concept alignment module} during the learning of the Concept Bottleneck Layer (CBL) to enhance neuron interpretability. This alignment enforces neurons to activate in response to target concepts, resulting in clearer concept representation while simultaneously improving predictive accuracy. Together, these modules aim to preserve past knowledge and improve both interpretability and performance in continual learning scenarios.

As shown in Figure~\ref{fig:framework}, we use the frozen CLIP image encoder $f_I$ and text encoder $f_T$ to extract image and concept features. When a new task arrives, we generate key concepts for each class using ChatGPT, selecting the most expressive and informative concepts  $\mathcal{C}_t$ to form a task-specific ``concept bottleneck'' using a learning-to-search algorithm, as proposed in \cite{LM4CV}. We save the bottlenecks that learned at each task for continual interpretability. We then construct the Concept Bottleneck Layer (CBL) to represent the relationships between learned classes and concepts, using both the current features and generated pseudo features of old classes. Inspired by \cite{LFCBM}, the concept score matrix output by the CBL is aligned with the CLIP concept activation matrix via a similarity loss $\mathcal{L}_{\text{sim}}$. Finally, the classifier maps these concept scores to predicted classes using cross-entropy loss $\mathcal{L}_{\text{ce}}$ and a sparsity loss $\mathcal{L}_{\text{sparse}}$, inspired by~\cite{sparseloss}, to encourage learning of key features and interpretable classifier.

\subsection{Model Training} 
\minisection{Concept Bottleneck Layer Construction.} We construct the Concept Bottleneck Layer (CBL) using  the bottlenecks collected 
across tasks to explain our model's behavior while enhancing its performance. The bottlenecks from tasks $\mathcal{T}_{1:t}$ are represented as $B_t = [z_1, z_2, \dots, z_k] \in \mathbb{R}^{|\mathcal{C}_{1:t}| \times D} $, $k=|\mathcal{C}_{1:t}|$,  where $D$ refers to the dimension of feature embedding. We save all previously learned bottlenecks, creating a cumulative concept space that preserves essential semantic knowledge to mitigate catastrophic forgetting. These bottlenecks also improve generalization to new tasks by leveraging a broader set of interpretable features accumulated across tasks. We use the frozen CLIP image encoder $f_I$ to extract image features $f_I(\mathcal{X}_t)$ and project these into the CBL to produce concept scores $E$, capturing the model’s understanding of relationships between images and concepts, as formulated below:
\begin{equation}
        E = f_I(\mathcal{X}_t) \cdot W_C^\top, \  E \in \mathbb{R}^{N_t \times |\mathcal{C}_{1:t}|}
\end{equation}
where $W_C$ represents the weight matrix of CBL, which consists of a single linear layer. Finally, the classifier $g$ maps $E$ to the class space, yielding the predicted labels $\hat{y}$ for the images.

Since the concepts are acquired incrementally under CIL scenario, both the CBL and the classifier need to expand to accommodate the learning of new concepts and categories, ensuring that they accurately map the embeddings while retaining the learned knowledge from the previous task. 

\minisection{Concept Alignment.} \label{mini:CA}
To guide the CBL toward human-understandable bottlenecks and enhance interpretability, we align the learned concept scores with CLIP concept activation scores. This alignment maintains semantic consistency across tasks and improves robustness by anchoring learning to stable, meaningful concepts.
By using CLIP, we can easily acquire the CLIP concept activation matrix $E_{clip}$ by equation \eqref{eq:clip_activation_matrix}. We train CBL to align with the CLIP concept score matrix $E_{clip}$, transferring the general knowledge to the output $E$ of CBL by a similarity loss $\mathcal{L}_{sim}$, defined as:
\begin{equation}
    \begin{aligned}
    \mathcal{L}_{sim} = & -\frac{1}{|\mathcal{C}_{1:t}|} \sum_{i=1}^{|\mathcal{C}_{1:t}|} \cos(\hat{E}^i, \hat{E}_{clip}^i) \\ 
                      = & -\frac{1}{|\mathcal{C}_{1:t}|} \sum_{i=1}^{|\mathcal{C}_{1:t}|} \frac{\hat{E}^i \cdot \hat{E}_{clip}^i}{||\hat{E}^i||_2 \cdot||\hat{E}_{clip}^i||_2}
    \end{aligned}
\end{equation}
Here, $\hat{E}=E^3$ and $\hat{E}_{clip}=E_{clip}^3$ to further sharp the concept scores distribution following~\cite{LFCBM}.

The classifier in CBM connects the concept space and the class space, can be understood as the contribution of concepts to distinguish categories. To encourage more interpretable classifier and transparent decision-making process, we regularize the final layer of our model using elastic-net penalty proposed by \cite{sparseloss} as shown below:
\begin{equation}
    \mathcal{L}_{sparse} =  \phi ||W_l||_1 + \frac{1}{2} (1-\phi) ||W_l||_F
\end{equation}
where $W_l$ is the weight of final layer, $\phi$ is a trade-off parameter, we set $\phi$ to 0.99 for our method.

\begin{table*}[!h]
\centering
\caption{Performance comparison on three coarse-grained datasets, the best performance is shown in bold, the second-best performance is underlined. All methods are implemented without using exemplars. We replace the backbones of all methods to CLIP ViT-B/16.}
\label{tab:Main results} 
\begin{adjustbox}{width=\textwidth,keepaspectratio}
\begin{tabular}{c cccc | cccc | cccc }
\toprule

\multirow{3}{*}{\textbf{Methods}} & \multicolumn{4}{c}{\textbf{CIFAR-100}} &  \multicolumn{4}{c}{\textbf{Tiny-ImageNet}} & \multicolumn{4}{c}{\textbf{ImageNet-subset}} \\

& \multicolumn{2}{c}{B-10 Inc-10} & \multicolumn{2}{c}{B-50 Inc-5} & \multicolumn{2}{c}{B-10 Inc-10} & \multicolumn{2}{c}{B-100 Inc-10} & \multicolumn{2}{c}{B-10 Inc-10} & \multicolumn{2}{c}{B-50 Inc-5} \\
& $\bar{A}$ & $A_{last}$ & $\bar{A}$ & $A_{last}$ & $\bar{A}$ & $A_{last}$ & $\bar{A}$ & $A_{last}$ & $\bar{A}$ & $A_{last}$ & $\bar{A}$ & $A_{last}$ \\
\midrule

\text{L2P\cite{l2p}} & 76.56 & 65.75 & 61.85 & 44.19 & 69.66 & 60.36 & 62.29 & 55.13 & 71.28 & 52.24 & 65.89 & 49.08 \\
\text{DualPrompt\cite{dualprompt}} & 81.41 & 70.34 & 64.05 & 43.86 & 74.06 & 66.08 & 65.35 & 56.15 & 72.86 & 54.20 & 64.64 & 49.48 \\
\text{CODA-Prompt\cite{codaprompt}} & 82.13 & 72.34 & 65.49 & 49.72 & 75.18 & 66.65 & 61.47 & 48.31 & 71.17 & 52.98 & 64.64 & 41.40\\
\text{CPP\cite{CPP}} & 75.73 & 67.50 & 69.57 & 66.26 & 68.70 & 61.23 & 66.61 & 63.48 & 83.45 & \underline{75.80} & 79.74 & 73.78 \\
\text{LAE\cite{LAE}} & 82.65 & 72.60 & 67.35 & 50.96 & 76.81 & 68.98 & 63.38 & 49.52 & 78.29 & 62.94 & 64.25 & 47.90\\
\text{Continual-CLIP\cite{con-clip}} & 75.15 & 66.68 & 70.79 & 66.68 & 63.63 & 55.91 & 58.68 & 55.91 & \underline{84.98} & 75.40 & \underline{81.35} & \underline{75.40} \\
\text{SLCA\cite{zhang2023slca}} & 83.13 & 72.01 & \textbf{80.07} & \underline{71.24} & 68.99 & 54.50 & 60.34 & 48.53 & 83.19 & 69.44 & 80.81 & 73.78 \\
\text{EASE\cite{EASE}} & \textbf{85.07} & \textbf{77.31} & 76.72 & 70.50 & \textbf{79.88} & \textbf{72.99} & \underline{70.42} & \underline{64.12} & 84.80 & 70.82 & 63.74 & 56.48 \\
\rowcolor{LightPurple}
\textbf{Ours} & \underline{84.49} \scriptsize{$\pm$ 0.26} & \underline{76.82} \scriptsize{$\pm$ 0.50} & \underline{79.07} \scriptsize{$\pm$ 0.34} & \textbf{75.91} \scriptsize{$\pm$ 0.50} & \underline{79.28} \scriptsize{$\pm$ 0.86} & \underline{71.98} \scriptsize{$\pm$ 0.25} & \textbf{75.64} \scriptsize{$\pm$ 0.17} & \textbf{71.97} \scriptsize{$\pm$ 0.09} & \textbf{86.83} \scriptsize{$\pm$ 1.32} & \textbf{78.97} \scriptsize{$\pm$ 0.39} & \textbf{81.85} \scriptsize{$\pm$ 0.76} & \textbf{78.21} \scriptsize{$\pm$ 0.29} \\

\bottomrule
\end{tabular}
\end{adjustbox}
\vspace{-2mm}
\end{table*}

\begin{table*}[!h]

\centering
\caption{Average incremental accuracy comparison on four fine-grained datasets, the best performance is shown in bold, the second-best is underlined. All methods are implemented without using exemplars. We replace the backbones of all methods to CLIP ViT-B/16.}
\label{tab:fine-grained results} 
\begin{adjustbox}{width=\textwidth,keepaspectratio}
\begin{tabular}{c cc | cc | cc | cc}
\toprule

\multirow{2}{*}{\textbf{Methods}}  & \multicolumn{2}{c}{\textbf{CUB-200}} & \multicolumn{2}{c}{\textbf{Flower}}  & \multicolumn{2}{c}{\textbf{Stanford-cars}} &  \multicolumn{2}{c}{\textbf{Food-101}}  \\

& B-10 Inc-10 & B-100 Inc-10 & B-10 Inc-10 & B-50 Inc-5 & B-14 Inc-14 & B-100 Inc-10 & B-10 Inc-10 & B-50 Inc-5 \\
\midrule
\text{L2P\cite{l2p}} & 62.08 & 59.38 & 84.37 & 76.70 & 64.42 & 61.82 & 79.48 & 69.72 \\
\text{DualPrompt\cite{dualprompt}} & 64.95 & 61.85 & 89.64 & 79.16 & 76.94 & 68.46 & 86.27 & 69.66 \\
\text{CODA-Prompt\cite{codaprompt}} & 67.22 & 59.82 & 88.57 & 77.70 & 76.44 & 60.80 & 87.76 & 67.22 \\
\text{CPP\cite{CPP}} & 83.60 & 75.03 & \underline{94.95} & \underline{93.30} & 84.75 & 77.49 & 90.21 & 86.60 \\
\text{LAE\cite{LAE}} & 66.45 & 59.98 & 86.79 & 77.55 & 77.25 & \underline{80.28} & 88.41 &  66.26 \\
\text{Continual-CLIP}\cite{con-clip} & 69.41 & 60.35 & 78.72 & 74.06 & \underline{86.43} & 69.79 & \underline{92.04} & \underline{89.96} \\
\text{SLCA\cite{zhang2023slca}} & 80.53 & \underline{76.85} & 92.77 & 82.14 & 84.74 & 70.59 & 74.49 & 76.28 \\
\text{EASE\cite{EASE}} & \underline{83.87} & 66.14 & 94.86 & 77.05 & 86.22 & 64.32 & 91.74 &  81.72 \\
\rowcolor{LightPurple}
\textbf{Ours} & \textbf{85.40} \scriptsize{$\pm$ 0.61} & \textbf{82.20} \scriptsize{$\pm$ 0.65} & \textbf{95.58} \scriptsize{$\pm$ 0.20} & \textbf{94.53} \scriptsize{$\pm$ 0.20} & \textbf{88.60} \scriptsize{$\pm$ 0.55} & \textbf{85.07} \scriptsize{$\pm$ 0.82} & \textbf{92.25} \scriptsize{$\pm$ 0.32} & \textbf{90.97} \scriptsize{$\pm$ 0.42} \\
\bottomrule
\end{tabular}
\end{adjustbox}
\vspace{-2mm}
\end{table*}

\minisection{Semantic-guided Prototype Augmentation.}
Prototype augmentation or compensation is commonly employed in class-incremental learning (CIL) to retain previously acquired knowledge, as prototypes from earlier tasks may become inaccurate over time due to feature drift. Research in \cite{SDC, fetril, PASS} demonstrates that feature drift within feature extractor can be estimated using data from the current task, resulting in significant performance gains. Building on this insight, we propose leveraging semantic knowledge to augment prototypes, thereby mitigating catastrophic forgetting and align concepts across all categories. Specifically, we propose a semantic-guided prototype augmentation strategy to generate pseudo-features for previously learned classes. We assume that classes with similar semantic information exhibit similar embedding distributions. Therefore, we first identify the category most similar to a previous class \( j \in \mathcal{Y}_{1:t-1}\) by calculating similarity scores between the embedding of the old class name and the prototypes of newly learned classes as follows:
\begin{equation}
    h = \operatorname*{arg\,max}_{i \in \mathcal{Y}_t} (\cos (f_T(y^j), p^i))
    \label{eq:semantic relation}
\end{equation}
where $y^j$ refers to the class name of class \textit{j}, $p_i$ refers to the prototype of class \textit{i}. 

Then we compute the discrepancy between the image features $V^h$ of class $h$ in the current task $t$ and the prototype of class $h$. We add this discrepancy to the prototype of class $j$ to obtain the generated pseudo features $\tilde{V}^j$, where $j$ is the class most semantically similar to $h$. The generated pseudo features $\tilde{V^j}$ of previous class $j$ can formulates as follows:
\begin{equation}
    \begin{aligned}
        \tilde V^j =p^j + & V^h - p^h
    \end{aligned}
    \label{eq:augment prototypes}
\end{equation}

After augmenting prototypes, we train our model on the current task data $\mathcal{D}_t$ along with the generated pseudo features guiding the learning process with cross-entropy loss as follows:
\begin{equation}
    \hat{y} = V_t \cdot W_C^\top \cdot W_l^\top
\end{equation}
where $V_t$ represents the feature set which involves the image features of classes in $\mathcal{Y}_t$ and augmented prototypes of classes in $\mathcal{Y}_{1:t-1}$. Overall, the final loss function can be described as:
\begin{equation}
    \begin{aligned}
        \mathcal{L} =& \mathcal{L}_{ce}(\hat{y}, y ) + \lambda \mathcal{L}_{sim} + \sigma \mathcal{L}_{sparse}
    \end{aligned}
    \label{eq:final loss}
\end{equation}

where $y \in \mathcal{Y}_{1:t}$, $\lambda$ and $\sigma$ are trade-off weights for $\mathcal{L}_{sim}$ and $\mathcal{L}_{sparse}$ respectively. 

\section{Experiments} \label{sec:exps}

\subsection{Experiments Setup}
\minisection{Evaluation Benchmarks.} 
We evaluated our method on three coarse-grained datasets: CIFAR-100~\cite{CIFAR100}, Tiny-ImageNet~\cite{Tiny}, ImageNet-subset~\cite{imagenet}. The first two datasets contain 100 classes each, while the latter contains 200 classes. In addition, we performed comprehensive evaluations on four fine-grained datasets: CUB-200~\cite{CUB200}, Flower~\cite{Flower}, Food-101~\cite{Food}, Stanford-cars~\cite{Cars}. There are 200 classes in CUB-200, 101 classes in Food-101, 102 classes in Flower, 196 classes in Stanford-cars.

We adopt the exemplar-free class incremental learning

setting, where no exemplars are retained for previously learned classes. The dataset is split according to the format “B-\textit{m} Inc-\textit{n}”, where \textit{m} denotes the number of classes included in the initial task, and \textit{n} represents the number of classes included in each incremental task. For each dataset, we perform experiments using two different splitting strategies: (1) large \textit{m} with small \textit{n}, and (2) \textit{m} = \textit{n}. All experiments we conducted are based on same random seed 1993 for fair comparison.

\minisection{Implementation Details.}
We developed our method and reproduced other method with Pilot \cite{pilot}. All experiments were conducted on an NVIDIA RTX 3090. Unless otherwise specified, we ran all experiments using CLIP ViT-B/16 as the backbone for all methods. CLIP RN-50 was used as backbone when comparing to ICICLE, which is designed for CNN-based model. We train our model using the Adam optimizer \cite{adam} with a batch size of 64, a learning rate of 0.001, and a total of 60 training epochs. We set the trade-off weight of similarity loss $\lambda$ to 1 and that of sparsity loss $\sigma$ to 0.001. The prompts we used to query ChatGPT and examples of generated concepts are provided in Supp. Mat. \ref{sec: query prompts}. More implementation details can be found in Supp. Mat. \ref{sec:Additional imp_details}, including the implementation of the concept selection module and the pseudo code of the algorithm.

\minisection{Evaluation Metric.}
Following the evaluation protocol of previous works~\cite{l2p,zhang2023slca,EASE}, we report the average incremental accuracy, denoted as $\bar{A} = \frac{1}{n}\sum_1^n A_t$, where $A_t$ represents the mean accuracy on all learned categories after learning task \textit{t}, $n$ denotes the number of tasks. Additionally, we also use $A_{last}$ to represent the average accuracy of all categories after learning the last task $n$.

\subsection{Comparison with State-of-the-Art Methods}
\minisection{Evaluation on Coarse-grained Datasets.}
In \Cref{tab:Main results}, we report the average incremental accuracy $\bar{A}$ and the final average accuracy $A_{last}$ of our method compared with state-of-the-art methods on three datasets: CIFAR-100, Tiny-ImageNet, ImageNet-subset. As shown, our approach demonstrates either the best or second-best performance across all settings and datasets. On ImageNet-subset, we achieve at least a 1.79\% improvement on \( \bar{A} \) and 3.06\% on \( A_{last} \) compared to existing methods. Our results on CIFAR-100 and Tiny-ImageNet are comparable to those methods that prioritize performance improvement without considering interpretability. 
In contrast, our method achieves a balanced approach, enhancing both performance and interpretability.

\minisection{Evaluation on Fine-grained Datasets.}
We also conduct experiments on four fine-grained datasets: CUB-200, Flowers, Stanford Cars, and Food-101. As shown in Table~\ref{tab:fine-grained results}, it is clear that our method achieves superior results across various data splits and datasets, particularly on CUB-200 and Stanford Cars, where the gains reach 4.43\% and 4.69\%, respectively. This suggests that our language-guided CBM-based approach provides significant benefits for fine-grained dataset classification in a continual learning setting.

\minisection{Comparison with Interpretable Methods.}
We further compared our proposed method with ICICLE\cite{ICICLE}, a recently proposed interpretability-driven continual learning method. The original ICICLE utilizes a ResNet pre-trained on ImageNet as its backbone. To ensure a fair comparison, we replace the backbone in both ICICLE and our method with the pre-trained CLIP RN50.
As shown in \Cref{tab:ICICLE ablation results}, we can observe that our method significantly outperforms ICICLE on both benchmark datasets and demonstrates greater stability to various data split. 

Moreover, our method requires fewer trainable parameters than ICICLE, which necessitates training the entire model. Additionally, our method offers greater flexibility, as it is architecture-agnostic. Unlike ICICLE, a prototypical-part-based approach constrained to CNN architectures, our method can utilize either a Vision Transformer \cite{vit} or ResNet as the backbone. 

Additional experimental results are provided in Supp. Mat. \ref{sec: more exps}, including dynamic performance curves under different settings, results obtained using various versions of description generation models, comparisons with joint training, and ablation studies on different modules.

\begin{table}[t]
\centering
\caption{Average incremental accuracy comparison with ICICLE, the best performance is shown in bold. We replace the backbone of our method and ICICLE to pre-trained CLIP RN50.}
\label{tab:ICICLE ablation results} 
\begin{adjustbox}{width=0.48\textwidth}
\begin{tabular}{c cc cc}
\toprule
\multirow{2}{*}{\textbf{Methods}}  & \multicolumn{2}{c}{\textbf{CUB-200}} & \multicolumn{2}{c}{\textbf{Stanford-cars}} \\
& B-50 Inc-50 & B-20 Inc-20 & B-49 Inc-49 & B-14 Inc-14\\
\midrule
\text{ICICLE~\cite{ICICLE}} & 39.31 & 20.85 & 49.58 & 24.22 \\
\textbf{Ours} & \textbf{66.21} & \textbf{69.72} & \textbf{73.91} & \textbf{75.26}\\
\bottomrule
\end{tabular}
\end{adjustbox}
\vspace{-2mm}
\end{table}

\subsection{Further Analysis}

\begin{figure}[t]
    \centering
    \includegraphics[width=0.45\textwidth]{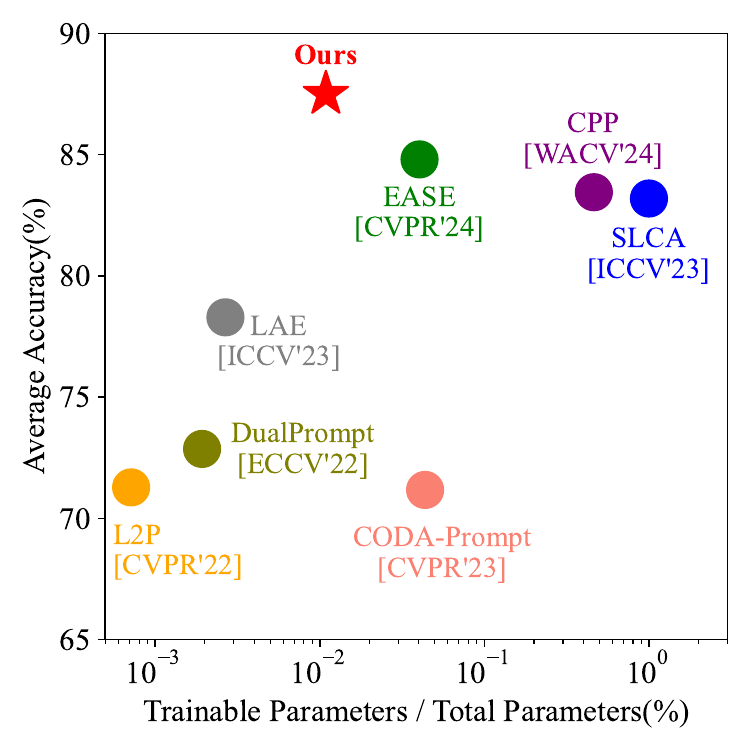}
    \caption{Performance-Parameter comparison of our method and benchmark methods on ImageNet-subset B-10 Inc-10.}
    \label{fig:AP_IN100}
\vspace{-2mm}
\end{figure}

\begin{figure*}[htbp]
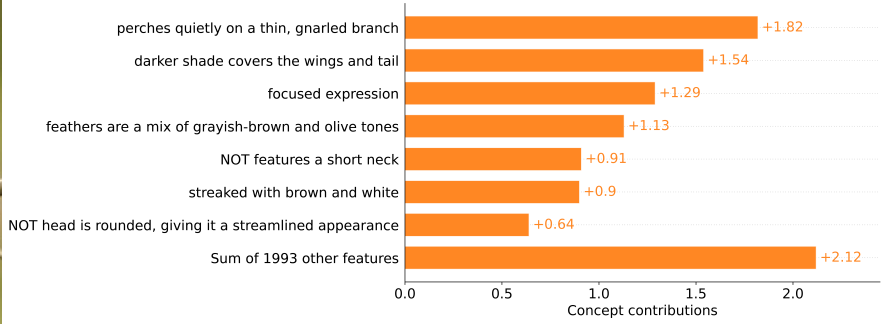

    \centering
    \begin{subfigure}[t]{\textwidth}
        \centering
        \includegraphics[width=0.95\textwidth]{final_weight_visual/CV_IN100-4.pdf}
        \caption{Contribution visualization on ImageNet-subset}
        \label{fig:CV-IN100}
    \end{subfigure}

    \vspace{10pt}

    \begin{subfigure}[t]{\textwidth}
        \centering
        \includegraphics[width=0.95\textwidth]{final_weight_visual/CV_CUB-6.pdf}
        \caption{Contribution visualization on CUB-200}
        \label{fig:CV-CUB}
    \end{subfigure}

    \caption{(a) Contribution visualization of \textit{axolotl} after training on ImageNet-subset B-10 Inc-10. (b) Contribution visualization of \textit{sayornis} after training on CUB-200 B-10 B-10.} 
    \label{fig:CV}
\vspace{-2mm}
\end{figure*}

\minisection{Performance-Parameter Comparison.}
\Cref{fig:AP_IN100} demonstrates the Performance-Parameter comparison between our method and other state-of-the-art methods after training on ImageNet-subset B-10 Inc-10. Methods with fewer trainable parameters and higher accuracy are considered superior. From \Cref{fig:AP_IN100} we can observe that all the compared methods, our method achieves best performance with relatively less trainable parameters. Since our model is designed to select concepts from a concept pool, approximately half of its trainable parameters are allocated to learning to identify expressive and informative concepts. The parameter requirement could be further reduced by using pre-defined concepts instead of training concept selection for each task. We also provide the computational overhead of the concept selection module in Supp. Mat. \ref{sec: CS cost}.

\minisection{Effect of Concept Quantity.} \Cref{tab:concept num} manifests the effect of different concept quantities to the performance. 
We report $\bar{A}$ of our method with 10,20,50 and 100 concepts per task 
on CUB-200 B-10 Inc-10 and ImageNet-subset B-10 Inc-10. We observe that performance improves as the number of concepts learned increases. On CUB-200 dataset, performance shows a dramatic improvement with more concepts per task, eventually plateauing. For ImageNet-subset, the results remain relatively stable as the number of concepts increases. Based on these observations, we choose to adapt 100 concepts per task across all datasets, striking a balance between concept diversity and computational efficiency.

\begin{figure*}
    \centering
    \includegraphics[width=\linewidth]{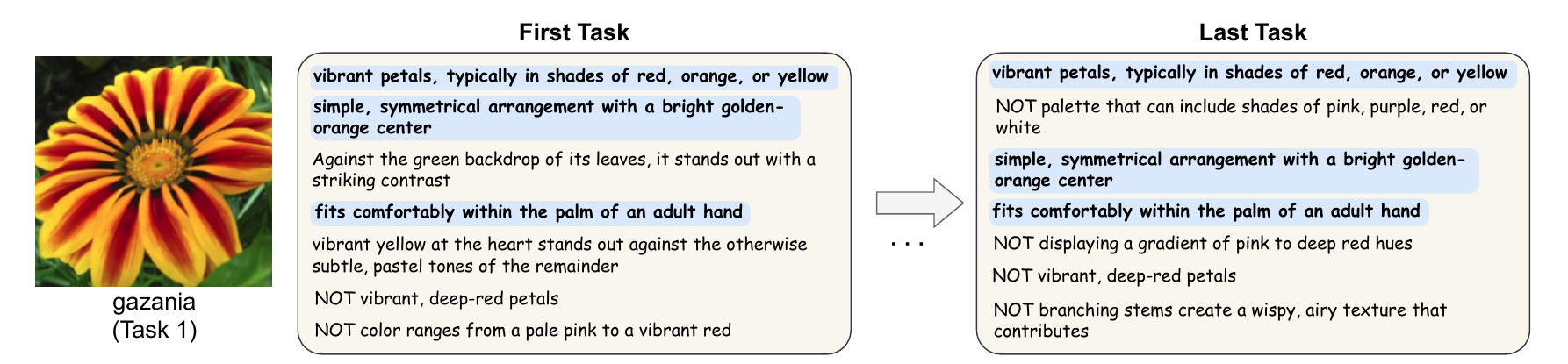}
    \caption{Top-7 concepts change of \textit{gazania} after training the first and all the tasks on Flower B-25 Inc-25.}
    \label{fig:cpt_drift}
    \vspace{-2mm}
\end{figure*}

\subsection{Understanding Model Interpretability}

\minisection{Interpretable Model Prediction.}
Inspired by~\cite{LFCBM,LM4CV,DNCBM}, We analyze the decision-making process of our method by highlighting the contribution of each concept to the final classification result.
Given an input image $x$ of class $k$, we denote the contribution of concept \( c_i \) to class \( k \) as \( Con_{i,k} \), which is computed as follows:
\begin{equation} 
    Con_{i,k} = f_I(x) \cdot (W_C^i)^\top \cdot W_l^{k,i}
\end{equation}
Here, $W_C^i$ denotes the $i$-th row of the weight matrix of the CBL, and $W_l^{k,i}$ represents the connection weight between class $k$ and concept $i$ in the final layer. The value of $Con_{i,k}$ can be either positive or negative, indicating positive or negative relevance to class $k$, respectively. A higher absolute value of $Con_{i,k}$ signifies greater importance of the concept for that class.

As illustrated in \Cref{fig:CV}, we present the top-7 most relevant concepts for an \textit{axolotl} image from ImageNet-subset and a \textit{sayornis} image from CUB-200 after the entire training process. Concepts are listed in descending order of absolute contribution,
with concepts labeled as ``NOT" indicating a negative association to the image. 
We can see that the concept ``speckled with brown and white” contributes most to the class \textit{axolotl} as shown in \Cref{fig:CV-IN100}, and ``perches quietly on a thin, gnarled branch” is highly activated to class \textit{sayornis} in \Cref{fig:CV-CUB}, the decision-making process is illustrated through human-interpretable concepts, providing insights into the model's reasoning. Furthermore, lower relevant concepts have minimal impact, as demonstrated by the cumulative negative contribution of 993 other features in \Cref{fig:CV-IN100}, indicating a weak association with \textit{axolotl}. Although the cumulative contribution of 1993 additional features in \Cref{fig:CV-CUB} exceeds that of the top-7 concepts, the contribution of each individual feature within the 1993 concepts is marginal. Notably, negative concepts also play a role in enhancing model performance and confidence, reflecting their significance in the interpretability framework. More examples of interpretable model predictions are provided in Supp. Mat. \ref{sec: more interpretable preds}.

\begin{table}[t]
    \centering
    \caption{The effect of the number of concepts on the performance of the method.}
    \begin{adjustbox}{width=0.4\textwidth}
    \begin{tabular}{c c c}
    \toprule
         \multirow{2}{*}{\textbf{Concepts per task}} & \multicolumn{1}{c}{\textbf{CUB-200}} & \multicolumn{1}{c}{\textbf{ImageNet-subset}}  \\
         & B-10 Inc-10 & B-10 Inc-10 \\
    \midrule
        10 & 79.32 & 86.25 \\
        20 & 83.70 & 86.80 \\
        50 & 85.70 & 87.24 \\
        100 & \textbf{86.07} & \textbf{87.40} \\
    \bottomrule
    \end{tabular}
    \end{adjustbox}
    \label{tab:concept num}

\vspace{-2mm}
\end{table}

\minisection{Concepts Change Across Tasks.}
As illustrated in \Cref{fig:cpt_drift}, we present the variation in concept relevance for the class \textit{gazania} on the Flower B-25 Inc-25, category \textit{gazania} is learned at first task. Concepts are ordered from highest to lowest based on their contribution values. Our method demonstrates robust interpretability, as concepts with high contribution scores in the initial task maintain relatively high contributions even after the completion of the final task. Additionally, the negative concepts also show increased contribution compared to the initial task, thereby alleviating forgetting of previous knowledge. It indicates that our framework help alleviate catastrophic forgetting by focusing on stable concepts that are shared across tasks, decoupling task-specific information, and promoting the retention of useful knowledge over time.

\section{Conclusion}
In this paper, we introduce a novel framework based on Language-Guided Concept Bottleneck Models for interpretable continual learning. Our approach enhances interpretability by aligning the concept score matrix generated by the concept bottleneck layer to the CLIP activation score matrix, while also learning a sparse linear classifier. Additionally, we propose a semantic-guided prototype augmentation to generate pseudo features for the previous tasks. Extensive experiments validate the effectiveness of our method, which maintains interpretability throughout the continual learning process.

\minisection{Limitation and Future Work.} 
Our approach builds upon Concept Bottleneck Models, making the model’s interpretability inherently dependent on the quality of the selected concepts. We have observed that the concepts generated by LLMs sometimes include non-visual descriptions, despite the use of appearance-related prompts. 
Future work will focus on improving concept selection quality and enhancing training efficiency.

{
    \small
    \bibliographystyle{ieeenat_fullname}
    \bibliography{main}

\begin{thebibliography}{72}
\providecommand{\natexlab}[1]{#1}
\providecommand{\url}[1]{\texttt{#1}}
\expandafter\ifx\csname urlstyle\endcsname\relax
  \providecommand{\doi}[1]{doi: #1}\else
  \providecommand{\doi}{doi: \begingroup \urlstyle{rm}\Url}\fi

\bibitem[Abbasnejad et~al.(2020)Abbasnejad, Teney, Parvaneh, Shi, and Hengel]{abbasnejad2020counterfactual}
Ehsan Abbasnejad, Damien Teney, Amin Parvaneh, Javen Shi, and Anton van~den Hengel.
\newblock Counterfactual vision and language learning.
\newblock In \emph{Proceedings of the IEEE/CVF conference on computer vision and pattern recognition}, pages 10044--10054, 2020.

\bibitem[Aljundi et~al.(2018)Aljundi, Babiloni, Elhoseiny, Rohrbach, and Tuytelaars]{aljundi2018memory}
Rahaf Aljundi, Francesca Babiloni, Mohamed Elhoseiny, Marcus Rohrbach, and Tinne Tuytelaars.
\newblock Memory aware synapses: Learning what (not) to forget.
\newblock In \emph{Proceedings of the European conference on computer vision (ECCV)}, pages 139--154, 2018.

\bibitem[Alvarez~Melis and Jaakkola(2018)]{alvarez2018towards}
David Alvarez~Melis and Tommi Jaakkola.
\newblock Towards robust interpretability with self-explaining neural networks.
\newblock \emph{Advances in neural information processing systems}, 31, 2018.

\bibitem[Bossard et~al.(2014)Bossard, Guillaumin, and Van~Gool]{Food}
Lukas Bossard, Matthieu Guillaumin, and Luc Van~Gool.
\newblock Food-101--mining discriminative components with random forests.
\newblock In \emph{Computer vision--ECCV 2014: 13th European conference, zurich, Switzerland, September 6-12, 2014, proceedings, part VI 13}, pages 446--461. Springer, 2014.

\bibitem[Brown et~al.(2020)Brown, Mann, Ryder, Subbiah, Kaplan, Dhariwal, Neelakantan, Shyam, Sastry, Askell, et~al.]{brown2020language}
Tom~B Brown, Benjamin Mann, Nick Ryder, Melanie Subbiah, Jared Kaplan, Prafulla Dhariwal, Arvind Neelakantan, Pranav Shyam, Girish Sastry, Amanda Askell, et~al.
\newblock Language models are few-shot learners.
\newblock \emph{Advances in Neural Information Processing Systems}, 33:\penalty0 1877--1901, 2020.

\bibitem[Chaudhry et~al.()Chaudhry, Ranzato, Rohrbach, and Elhoseiny]{chaudhryefficient}
Arslan Chaudhry, Marc’Aurelio Ranzato, Marcus Rohrbach, and Mohamed Elhoseiny.
\newblock Efficient lifelong learning with a-gem.
\newblock In \emph{International Conference on Learning Representations}.

\bibitem[Chen et~al.(2019)Chen, Li, Tao, Barnett, Rudin, and Su]{PPnet}
Chaofan Chen, Oscar Li, Daniel Tao, Alina Barnett, Cynthia Rudin, and Jonathan~K Su.
\newblock This looks like that: deep learning for interpretable image recognition.
\newblock \emph{Advances in neural information processing systems}, 32, 2019.

\bibitem[De~Lange et~al.(2021)De~Lange, Aljundi, Masana, Parisot, Jia, Leonardis, Slabaugh, and Tuytelaars]{de2021continual}
Matthias De~Lange, Rahaf Aljundi, Marc Masana, Sarah Parisot, Xu Jia, Ale{\v{s}} Leonardis, Gregory Slabaugh, and Tinne Tuytelaars.
\newblock A continual learning survey: Defying forgetting in classification tasks.
\newblock \emph{IEEE transactions on pattern analysis and machine intelligence}, 44\penalty0 (7):\penalty0 3366--3385, 2021.

\bibitem[Deng et~al.(2009)Deng, Dong, Socher, Li, Li, and Fei-Fei]{imagenet}
Jia Deng, Wei Dong, Richard Socher, Li-Jia Li, Kai Li, and Li Fei-Fei.
\newblock Imagenet: A large-scale hierarchical image database.
\newblock In \emph{2009 IEEE conference on computer vision and pattern recognition}, pages 248--255. Ieee, 2009.

\bibitem[Dhar et~al.(2019)Dhar, Singh, Peng, Wu, and Chellappa]{LWM}
Prithviraj Dhar, Rajat~Vikram Singh, Kuan-Chuan Peng, Ziyan Wu, and Rama Chellappa.
\newblock Learning without memorizing.
\newblock In \emph{Proceedings of the IEEE/CVF conference on computer vision and pattern recognition}, pages 5138--5146, 2019.

\bibitem[Donnelly et~al.(2022)Donnelly, Barnett, and Chen]{DPPnet}
Jon Donnelly, Alina~Jade Barnett, and Chaofan Chen.
\newblock Deformable protopnet: An interpretable image classifier using deformable prototypes.
\newblock In \emph{Proceedings of the IEEE/CVF conference on computer vision and pattern recognition}, pages 10265--10275, 2022.

\bibitem[Dosovitskiy(2020)]{vit}
Alexey Dosovitskiy.
\newblock An image is worth 16x16 words: Transformers for image recognition at scale.
\newblock \emph{arXiv preprint arXiv:2010.11929}, 2020.

\bibitem[Gao et~al.(2023)Gao, Zhao, Sun, Xi, Zhang, Ghanem, and Zhang]{LAE}
Qiankun Gao, Chen Zhao, Yifan Sun, Teng Xi, Gang Zhang, Bernard Ghanem, and Jian Zhang.
\newblock A unified continual learning framework with general parameter-efficient tuning.
\newblock In \emph{Proceedings of the IEEE/CVF International Conference on Computer Vision}, pages 11483--11493, 2023.

\bibitem[Gomez-Villa et~al.(2022)Gomez-Villa, Twardowski, Yu, Bagdanov, and van~de Weijer]{Gomez-Villa_2022_CVPR}
Alex Gomez-Villa, Bartlomiej Twardowski, Lu Yu, Andrew~D. Bagdanov, and Joost van~de Weijer.
\newblock Continually learning self-supervised representations with projected functional regularization.
\newblock In \emph{Proceedings of the IEEE/CVF Conference on Computer Vision and Pattern Recognition (CVPR) Workshops}, pages 3867--3877, 2022.

\bibitem[Goyal et~al.(2019)Goyal, Wu, Ernst, Batra, Parikh, and Lee]{goyal2019counterfactual}
Yash Goyal, Ziyan Wu, Jan Ernst, Dhruv Batra, Devi Parikh, and Stefan Lee.
\newblock Counterfactual visual explanations.
\newblock In \emph{International Conference on Machine Learning}, pages 2376--2384. PMLR, 2019.

\bibitem[Hou et~al.(2019)Hou, Pan, Loy, Wang, and Lin]{hou2019learning}
Saihui Hou, Xinyu Pan, Chen~Change Loy, Zilei Wang, and Dahua Lin.
\newblock Learning a unified classifier incrementally via rebalancing.
\newblock In \emph{Proceedings of the IEEE/CVF conference on computer vision and pattern recognition}, pages 831--839, 2019.

\bibitem[Huang et~al.(2024)Huang, Cao, Lu, and Liu]{RAPF}
Linlan Huang, Xusheng Cao, Haori Lu, and Xialei Liu.
\newblock Class-incremental learning with clip: Adaptive representation adjustment and parameter fusion.
\newblock In \emph{European Conference on Computer Vision}, pages 214--231. Springer, 2024.

\bibitem[Kim et~al.(2018)Kim, Wattenberg, Gilmer, Cai, Wexler, Viegas, et~al.]{kim2018interpretability}
Been Kim, Martin Wattenberg, Justin Gilmer, Carrie Cai, James Wexler, Fernanda Viegas, et~al.
\newblock Interpretability beyond feature attribution: Quantitative testing with concept activation vectors (tcav).
\newblock In \emph{International conference on machine learning}, pages 2668--2677. PMLR, 2018.

\bibitem[Kim et~al.(2022)Kim, Nam, and Ko]{kim2022vit}
Sangwon Kim, Jaeyeal Nam, and Byoung~Chul Ko.
\newblock Vit-net: Interpretable vision transformers with neural tree decoder.
\newblock In \emph{International conference on machine learning}, pages 11162--11172. PMLR, 2022.

\bibitem[Kingma(2014)]{adam}
Diederik~P Kingma.
\newblock Adam: A method for stochastic optimization.
\newblock \emph{arXiv preprint arXiv:1412.6980}, 2014.

\bibitem[Kirkpatrick et~al.(2017)Kirkpatrick, Pascanu, Rabinowitz, Veness, Desjardins, Rusu, Milan, Quan, Ramalho, Grabska-Barwinska, et~al.]{EWC}
James Kirkpatrick, Razvan Pascanu, Neil Rabinowitz, Joel Veness, Guillaume Desjardins, Andrei~A Rusu, Kieran Milan, John Quan, Tiago Ramalho, Agnieszka Grabska-Barwinska, et~al.
\newblock Overcoming catastrophic forgetting in neural networks.
\newblock \emph{Proceedings of the national academy of sciences}, 114\penalty0 (13):\penalty0 3521--3526, 2017.

\bibitem[Koh et~al.(2020)Koh, Nguyen, Tang, Mussmann, Pierson, Kim, and Liang]{CBM}
Pang~Wei Koh, Thao Nguyen, Yew~Siang Tang, Stephen Mussmann, Emma Pierson, Been Kim, and Percy Liang.
\newblock Concept bottleneck models.
\newblock In \emph{International conference on machine learning}, pages 5338--5348. PMLR, 2020.

\bibitem[Krause et~al.(2013)Krause, Stark, Deng, and Fei-Fei]{Cars}
Jonathan Krause, Michael Stark, Jia Deng, and Li Fei-Fei.
\newblock 3d object representations for fine-grained categorization.
\newblock In \emph{Proceedings of the IEEE international conference on computer vision workshops}, pages 554--561, 2013.

\bibitem[Krizhevsky et~al.(2009)Krizhevsky, Hinton, et~al.]{CIFAR100}
Alex Krizhevsky, Geoffrey Hinton, et~al.
\newblock Learning multiple layers of features from tiny images.
\newblock 2009.

\bibitem[Le and Yang(2015)]{Tiny}
Ya Le and Xuan Yang.
\newblock Tiny imagenet visual recognition challenge.
\newblock \emph{CS 231N}, 7\penalty0 (7):\penalty0 3, 2015.

\bibitem[Li and Hoiem(2017)]{lwf}
Zhizhong Li and Derek Hoiem.
\newblock Learning without forgetting.
\newblock \emph{IEEE transactions on pattern analysis and machine intelligence}, 40\penalty0 (12):\penalty0 2935--2947, 2017.

\bibitem[Li et~al.(2024)Li, Zhao, Zhang, Zhang, Liu, Liu, and Metaxas]{CPP}
Zhuowei Li, Long Zhao, Zizhao Zhang, Han Zhang, Di Liu, Ting Liu, and Dimitris~N Metaxas.
\newblock Steering prototypes with prompt-tuning for rehearsal-free continual learning.
\newblock In \emph{Proceedings of the IEEE/CVF Winter Conference on Applications of Computer Vision}, pages 2523--2533, 2024.

\bibitem[Liu et~al.(2024{\natexlab{a}})Liu, Diao, Huang, An, An, and Xu]{CLFD}
RuiQi Liu, Boyu Diao, Libo Huang, Zijia An, Zhulin An, and Yongjun Xu.
\newblock Continual learning in the frequency domain.
\newblock In \emph{Advances in Neural Information Processing Systems}, pages 85389--85411, 2024{\natexlab{a}}.

\bibitem[Liu et~al.(2024{\natexlab{b}})Liu, Zhai, Bagdanov, Li, and Cheng]{TASS}
Xialei Liu, Jiang-Tian Zhai, Andrew~D. Bagdanov, Ke Li, and Ming-Ming Cheng.
\newblock Task-adaptive saliency guidance for exemplar-free class incremental learning.
\newblock In \emph{Proceedings of the IEEE/CVF Conference on Computer Vision and Pattern Recognition (CVPR)}, pages 23954--23963, 2024{\natexlab{b}}.

\bibitem[Mallya and Lazebnik(2018)]{PACKnet}
Arun Mallya and Svetlana Lazebnik.
\newblock Packnet: Adding multiple tasks to a single network by iterative pruning.
\newblock In \emph{Proceedings of the IEEE conference on Computer Vision and Pattern Recognition}, pages 7765--7773, 2018.

\bibitem[Masana et~al.(2022)Masana, Liu, Twardowski, Menta, Bagdanov, and Van De~Weijer]{masana2022class}
Marc Masana, Xialei Liu, Bart{\l}omiej Twardowski, Mikel Menta, Andrew~D Bagdanov, and Joost Van De~Weijer.
\newblock Class-incremental learning: survey and performance evaluation on image classification.
\newblock \emph{IEEE Transactions on Pattern Analysis and Machine Intelligence}, 45\penalty0 (5):\penalty0 5513--5533, 2022.

\bibitem[McCloskey and Cohen(1989)]{mccloskey1989catastrophic}
Michael McCloskey and Neal~J Cohen.
\newblock Catastrophic interference in connectionist networks: The sequential learning problem.
\newblock In \emph{Psychology of learning and motivation}, pages 109--165. Elsevier, 1989.

\bibitem[Mothilal et~al.(2020)Mothilal, Sharma, and Tan]{mothilal2020explaining}
Ramaravind~K Mothilal, Amit Sharma, and Chenhao Tan.
\newblock Explaining machine learning classifiers through diverse counterfactual explanations.
\newblock In \emph{Proceedings of the 2020 conference on fairness, accountability, and transparency}, pages 607--617, 2020.

\bibitem[Nauta et~al.(2021)Nauta, Van~Bree, and Seifert]{nauta2021neural}
Meike Nauta, Ron Van~Bree, and Christin Seifert.
\newblock Neural prototype trees for interpretable fine-grained image recognition.
\newblock In \emph{Proceedings of the IEEE/CVF conference on computer vision and pattern recognition}, pages 14933--14943, 2021.

\bibitem[Nilsback and Zisserman(2008)]{Flower}
Maria-Elena Nilsback and Andrew Zisserman.
\newblock Automated flower classification over a large number of classes.
\newblock In \emph{2008 Sixth Indian conference on computer vision, graphics \& image processing}, pages 722--729. IEEE, 2008.

\bibitem[Oikarinen et~al.(2023)Oikarinen, Das, Nguyen, and Weng]{LFCBM}
Tuomas Oikarinen, Subhro Das, Lam Nguyen, and Lily Weng.
\newblock Label-free concept bottleneck models.
\newblock In \emph{International Conference on Learning Representations}, 2023.

\bibitem[Petit et~al.(2023)Petit, Popescu, Schindler, Picard, and Delezoide]{fetril}
Gr{\'e}goire Petit, Adrian Popescu, Hugo Schindler, David Picard, and Bertrand Delezoide.
\newblock Fetril: Feature translation for exemplar-free class-incremental learning.
\newblock In \emph{Proceedings of the IEEE/CVF winter conference on applications of computer vision}, pages 3911--3920, 2023.

\bibitem[Radford et~al.(2021)Radford, Kim, Hallacy, Ramesh, Goh, Agarwal, Sastry, Askell, Mishkin, Clark, et~al.]{CLIP}
Alec Radford, Jong~Wook Kim, Chris Hallacy, Aditya Ramesh, Gabriel Goh, Sandhini Agarwal, Girish Sastry, Amanda Askell, Pamela Mishkin, Jack Clark, et~al.
\newblock Learning transferable visual models from natural language supervision.
\newblock In \emph{International conference on machine learning}, pages 8748--8763. PMLR, 2021.

\bibitem[Rajasegaran et~al.(2019)Rajasegaran, Hayat, Khan, Khan, and Shao]{RPSnet}
Jathushan Rajasegaran, Munawar Hayat, Salman~H Khan, Fahad~Shahbaz Khan, and Ling Shao.
\newblock Random path selection for continual learning.
\newblock \emph{Advances in neural information processing systems}, 32, 2019.

\bibitem[Rao et~al.(2024)Rao, Mahajan, B{\"o}hle, and Schiele]{DNCBM}
Sukrut~Sridhar Rao, Sweta Mahajan, Moritz B{\"o}hle, and Bernt Schiele.
\newblock Discover-then-name: Task-agnostic concept bottlenecks via automated concept discovery.
\newblock In \emph{18th European Conference on Computer Vision}. Springer, 2024.

\bibitem[Rebuffi et~al.(2017)Rebuffi, Kolesnikov, Sperl, and Lampert]{icarl}
Sylvestre-Alvise Rebuffi, Alexander Kolesnikov, Georg Sperl, and Christoph~H Lampert.
\newblock icarl: Incremental classifier and representation learning.
\newblock In \emph{Proceedings of the IEEE conference on Computer Vision and Pattern Recognition}, pages 2001--2010, 2017.

\bibitem[Rolnick et~al.(2019)Rolnick, Ahuja, Schwarz, Lillicrap, and Wayne]{ER}
David Rolnick, Arun Ahuja, Jonathan Schwarz, Timothy Lillicrap, and Gregory Wayne.
\newblock Experience replay for continual learning.
\newblock \emph{Advances in neural information processing systems}, 32, 2019.

\bibitem[Rymarczyk et~al.(2021)Rymarczyk, Struski, Tabor, and Zieli{\'n}ski]{rymarczyk2021protopshare}
Dawid Rymarczyk, {\L}ukasz Struski, Jacek Tabor, and Bartosz Zieli{\'n}ski.
\newblock Protopshare: Prototypical parts sharing for similarity discovery in interpretable image classification.
\newblock In \emph{Proceedings of the 27th ACM SIGKDD Conference on Knowledge Discovery \& Data Mining}, pages 1420--1430, 2021.

\bibitem[Rymarczyk et~al.(2023)Rymarczyk, van~de Weijer, Zieli{\'n}ski, and Twardowski]{ICICLE}
Dawid Rymarczyk, Joost van~de Weijer, Bartosz Zieli{\'n}ski, and Bartlomiej Twardowski.
\newblock Icicle: Interpretable class incremental continual learning.
\newblock In \emph{Proceedings of the IEEE/CVF International Conference on Computer Vision}, pages 1887--1898, 2023.

\bibitem[Selvaraju et~al.(2017)Selvaraju, Cogswell, Das, Vedantam, Parikh, and Batra]{Gradcam}
Ramprasaath~R Selvaraju, Michael Cogswell, Abhishek Das, Ramakrishna Vedantam, Devi Parikh, and Dhruv Batra.
\newblock Grad-cam: Visual explanations from deep networks via gradient-based localization.
\newblock In \emph{Proceedings of the IEEE international conference on computer vision}, pages 618--626, 2017.

\bibitem[Selvaraju et~al.(2019)Selvaraju, Lee, Shen, Jin, Ghosh, Heck, Batra, and Parikh]{selvaraju2019taking}
Ramprasaath~R Selvaraju, Stefan Lee, Yilin Shen, Hongxia Jin, Shalini Ghosh, Larry Heck, Dhruv Batra, and Devi Parikh.
\newblock Taking a hint: Leveraging explanations to make vision and language models more grounded.
\newblock In \emph{Proceedings of the IEEE/CVF international conference on computer vision}, pages 2591--2600, 2019.

\bibitem[Shang et~al.(2024)Shang, Zhou, Zhang, Ni, Yang, and Wang]{shang2024incremental}
Chenming Shang, Shiji Zhou, Hengyuan Zhang, Xinzhe Ni, Yujiu Yang, and Yuwang Wang.
\newblock Incremental residual concept bottleneck models.
\newblock In \emph{Proceedings of the IEEE/CVF Conference on Computer Vision and Pattern Recognition}, pages 11030--11040, 2024.

\bibitem[Sheth and Ebrahimi~Kahou(2024)]{CoopCBM}
Ivaxi Sheth and Samira Ebrahimi~Kahou.
\newblock Auxiliary losses for learning generalizable concept-based models.
\newblock \emph{Advances in Neural Information Processing Systems}, 36, 2024.

\bibitem[Smith et~al.(2023)Smith, Karlinsky, Gutta, Cascante-Bonilla, Kim, Arbelle, Panda, Feris, and Kira]{codaprompt}
James~Seale Smith, Leonid Karlinsky, Vyshnavi Gutta, Paola Cascante-Bonilla, Donghyun Kim, Assaf Arbelle, Rameswar Panda, Rogerio Feris, and Zsolt Kira.
\newblock Coda-prompt: Continual decomposed attention-based prompting for rehearsal-free continual learning.
\newblock In \emph{Proceedings of the IEEE/CVF Conference on Computer Vision and Pattern Recognition}, pages 11909--11919, 2023.

\bibitem[Sun et~al.(2023)Sun, Zhou, Ye, and Zhan]{pilot}
Hai-Long Sun, Da-Wei Zhou, Han-Jia Ye, and De-Chuan Zhan.
\newblock Pilot: A pre-trained model-based continual learning toolbox.
\newblock \emph{arXiv preprint arXiv:2309.07117}, 2023.

\bibitem[Tao et~al.(2024)Tao, Yu, Yao, Huang, and Xu]{CIL4LWNetworks}
Zhe Tao, Lu Yu, Hantao Yao, Shucheng Huang, and Changsheng Xu.
\newblock Class incremental learning for light-weighted networks.
\newblock \emph{IEEE Transactions on Circuits and Systems for Video Technology}, 2024.

\bibitem[Thengane et~al.(2022)Thengane, Khan, Hayat, and Khan]{con-clip}
Vishal Thengane, Salman Khan, Munawar Hayat, and Fahad Khan.
\newblock Clip model is an efficient continual learner.
\newblock \emph{arXiv preprint arXiv:2210.03114}, 2022.

\bibitem[Wah et~al.(2011)Wah, Branson, Welinder, Perona, and Belongie]{CUB200}
Catherine Wah, Steve Branson, Peter Welinder, Pietro Perona, and Serge Belongie.
\newblock The caltech-ucsd birds-200-2011 dataset.
\newblock 2011.

\bibitem[Wang et~al.(2023)Wang, Li, Nakashima, and Nagahara]{wang2023learning}
Bowen Wang, Liangzhi Li, Yuta Nakashima, and Hajime Nagahara.
\newblock Learning bottleneck concepts in image classification.
\newblock In \emph{Proceedings of the ieee/cvf conference on computer vision and pattern recognition}, pages 10962--10971, 2023.

\bibitem[Wang et~al.(2022{\natexlab{a}})Wang, Zhang, Ebrahimi, Sun, Zhang, Lee, Ren, Su, Perot, Dy, et~al.]{dualprompt}
Zifeng Wang, Zizhao Zhang, Sayna Ebrahimi, Ruoxi Sun, Han Zhang, Chen-Yu Lee, Xiaoqi Ren, Guolong Su, Vincent Perot, Jennifer Dy, et~al.
\newblock Dualprompt: Complementary prompting for rehearsal-free continual learning.
\newblock In \emph{European Conference on Computer Vision}, pages 631--648, 2022{\natexlab{a}}.

\bibitem[Wang et~al.(2022{\natexlab{b}})Wang, Zhang, Lee, Zhang, Sun, Ren, Su, Perot, Dy, and Pfister]{l2p}
Zifeng Wang, Zizhao Zhang, Chen-Yu Lee, Han Zhang, Ruoxi Sun, Xiaoqi Ren, Guolong Su, Vincent Perot, Jennifer Dy, and Tomas Pfister.
\newblock Learning to prompt for continual learning.
\newblock In \emph{Proceedings of the IEEE/CVF conference on computer vision and pattern recognition}, pages 139--149, 2022{\natexlab{b}}.

\bibitem[Wong et~al.(2021)Wong, Santurkar, and Madry]{sparseloss}
Eric Wong, Shibani Santurkar, and Aleksander Madry.
\newblock Leveraging sparse linear layers for debuggable deep networks.
\newblock In \emph{International Conference on Machine Learning}, pages 11205--11216. PMLR, 2021.

\bibitem[Yan et~al.(2023)Yan, Wang, Zhong, Dong, He, Lu, Wang, Shang, and McAuley]{LM4CV}
An Yan, Yu Wang, Yiwu Zhong, Chengyu Dong, Zexue He, Yujie Lu, William~Yang Wang, Jingbo Shang, and Julian McAuley.
\newblock Learning concise and descriptive attributes for visual recognition.
\newblock In \emph{Proceedings of the IEEE/CVF International Conference on Computer Vision}, pages 3090--3100, 2023.

\bibitem[Yan et~al.(2021)Yan, Xie, and He]{DER}
Shipeng Yan, Jiangwei Xie, and Xuming He.
\newblock Der: Dynamically expandable representation for class incremental learning.
\newblock In \emph{Proceedings of the IEEE/CVF conference on computer vision and pattern recognition}, pages 3014--3023, 2021.

\bibitem[Yang et~al.(2023{\natexlab{a}})Yang, Cui, Xu, Zhong, Zheng, and Wang]{yang2023continual}
Yang Yang, Zhiying Cui, Junjie Xu, Changhong Zhong, Wei-Shi Zheng, and Ruixuan Wang.
\newblock Continual learning with bayesian model based on a fixed pre-trained feature extractor.
\newblock \emph{Visual Intelligence}, 1\penalty0 (1):\penalty0 5, 2023{\natexlab{a}}.

\bibitem[Yang et~al.(2023{\natexlab{b}})Yang, Panagopoulou, Zhou, Jin, Callison-Burch, and Yatskar]{LABO}
Yue Yang, Artemis Panagopoulou, Shenghao Zhou, Daniel Jin, Chris Callison-Burch, and Mark Yatskar.
\newblock Language in a bottle: Language model guided concept bottlenecks for interpretable image classification.
\newblock In \emph{Proceedings of the IEEE/CVF Conference on Computer Vision and Pattern Recognition}, pages 19187--19197, 2023{\natexlab{b}}.

\bibitem[Yeh et~al.(2020)Yeh, Kim, Arik, Li, Pfister, and Ravikumar]{yeh2020completeness}
Chih-Kuan Yeh, Been Kim, Sercan Arik, Chun-Liang Li, Tomas Pfister, and Pradeep Ravikumar.
\newblock On completeness-aware concept-based explanations in deep neural networks.
\newblock \emph{Advances in neural information processing systems}, 33:\penalty0 20554--20565, 2020.

\bibitem[Yoon et~al.(2018)Yoon, Yang, Lee, and Hwang]{yoon2018lifelong}
Jaehong Yoon, Eunho Yang, Jeongtae Lee, and Sung~Ju Hwang.
\newblock Lifelong learning with dynamically expandable networks.
\newblock In \emph{International Conference on Learning Representations}, 2018.

\bibitem[Yu et~al.(2020)Yu, Twardowski, Liu, Herranz, Wang, Cheng, Jui, and Weijer]{SDC}
Lu Yu, Bartlomiej Twardowski, Xialei Liu, Luis Herranz, Kai Wang, Yongmei Cheng, Shangling Jui, and Joost van~de Weijer.
\newblock Semantic drift compensation for class-incremental learning.
\newblock In \emph{Proceedings of the IEEE/CVF conference on computer vision and pattern recognition}, pages 6982--6991, 2020.

\bibitem[Yuksekgonul et~al.(2022)Yuksekgonul, Wang, and Zou]{PCBM}
Mert Yuksekgonul, Maggie Wang, and James Zou.
\newblock Post-hoc concept bottleneck models.
\newblock \emph{arXiv preprint arXiv:2205.15480}, 2022.

\bibitem[Zenke et~al.(2017)Zenke, Poole, and Ganguli]{zenke2017continual}
Friedemann Zenke, Ben Poole, and Surya Ganguli.
\newblock Continual learning through synaptic intelligence.
\newblock In \emph{International conference on machine learning}, pages 3987--3995. PMLR, 2017.

\bibitem[Zhai et~al.(2023)Zhai, Liu, Bagdanov, Li, and Cheng]{Bilateral_MAE}
Jiang-Tian Zhai, Xialei Liu, Andrew~D. Bagdanov, Ke Li, and Ming-Ming Cheng.
\newblock Masked autoencoders are efficient class incremental learners.
\newblock In \emph{Proceedings of the IEEE/CVF International Conference on Computer Vision (ICCV)}, pages 19104--19113, 2023.

\bibitem[Zhai et~al.(2024)Zhai, Liu, Yu, and Cheng]{zhai2024fine}
Jiang-Tian Zhai, Xialei Liu, Lu Yu, and Ming-Ming Cheng.
\newblock Fine-grained knowledge selection and restoration for non-exemplar class incremental learning.
\newblock In \emph{Proceedings of the AAAI Conference on Artificial Intelligence}, pages 6971--6978, 2024.

\bibitem[Zhang et~al.(2023)Zhang, Wang, Kang, Chen, and Wei]{zhang2023slca}
Gengwei Zhang, Liyuan Wang, Guoliang Kang, Ling Chen, and Yunchao Wei.
\newblock Slca: Slow learner with classifier alignment for continual learning on a pre-trained model.
\newblock In \emph{Proceedings of the IEEE/CVF International Conference on Computer Vision}, pages 19148--19158, 2023.

\bibitem[Zhou et~al.(2024{\natexlab{a}})Zhou, Sun, Ye, and Zhan]{EASE}
Da-Wei Zhou, Hai-Long Sun, Han-Jia Ye, and De-Chuan Zhan.
\newblock Expandable subspace ensemble for pre-trained model-based class-incremental learning.
\newblock In \emph{Proceedings of the IEEE/CVF Conference on Computer Vision and Pattern Recognition}, pages 23554--23564, 2024{\natexlab{a}}.

\bibitem[Zhou et~al.(2024{\natexlab{b}})Zhou, Wang, Qi, Ye, Zhan, and Liu]{zhou2024class}
Da-Wei Zhou, Qi-Wei Wang, Zhi-Hong Qi, Han-Jia Ye, De-Chuan Zhan, and Ziwei Liu.
\newblock Class-incremental learning: A survey.
\newblock \emph{IEEE Transactions on Pattern Analysis and Machine Intelligence}, 2024{\natexlab{b}}.

\bibitem[Zhu et~al.(2021)Zhu, Zhang, Wang, Yin, and Liu]{PASS}
Fei Zhu, Xu-Yao Zhang, Chuang Wang, Fei Yin, and Cheng-Lin Liu.
\newblock Prototype augmentation and self-supervision for incremental learning.
\newblock In \emph{Proceedings of the IEEE/CVF Conference on Computer Vision and Pattern Recognition}, pages 5871--5880, 2021.

\end{thebibliography}
}

\clearpage
\setcounter{page}{1}
\maketitlesupplementary

\section{Supplementary}
\subsection{Concepts Preparation} \label{sec: query prompts}
Before initiating the incremental learning process, we retrieve concepts associated with specific categories through ChatGPT queries. Following the methodologies proposed in \cite{LABO, LM4CV, LFCBM}, we employ the following prompts to guide the ChatGPT interactions:

\begin{table}[h]
\centering
\setlength{\abovecaptionskip}{1pt} 
\setlength{\belowcaptionskip}{-5pt} 
\captionsetup[table]{justification=centering, singlelinecheck=false} 
\caption{Prompts for all benchmark datasets.}
\begin{adjustbox}{width=0.5\textwidth}
{\fontsize{20}{27}\selectfont
\begin{tabular}{c l}
\toprule
\textbf{Dataset} & \textbf{Prompts}  \\
\midrule
\multirow{3}{*}{\parbox{5cm}{\centering Coarse-grained,  \\ CUB-200}} 
    & ``using \{num\} sentenses to discribe the\\ 
    & \textbf{ appearance / color / size / shape / surroundings}\\ 
    & of \{category\}" \\
\midrule
\multirow{3}{*}{Food101} 
    & ``using \{num\} sentences to describe the \\ 
    & \textbf{apperance / shape / color / texture} \\ 
    & of a food named \{category\}"\\
\midrule
\multirow{3}{*}{Flower} 
    & ``using \{num\} sentences to describe the \\ 
    & \textbf{apperance / color / size / patten / texture} \\ 
    & of a flower named \{category\}"\\
\midrule
\multirow{3}{*}{Stanford-cars} 
& ``using \{num\} sentences to describe the \\ 
& \textbf{appearance / shape / color / size / structure} \\ 
& of a car named \{category\}"\\
\bottomrule
\end{tabular}}
\label{tab:prompts} 
\end{adjustbox}
\end{table}

By utilizing ChatGPT with prompts mentioned above, we generate descriptive sentences containing class-specific concepts. Subsequently, we employ a T5 model, as redesigned by \cite{LABO}, to extract concepts from these sentences, thereby constructing a comprehensive general concept pool. Examples of extracted concepts for categories in CUB-200, ImageNet-subset, Food-101 and Flower are provided in \Cref{tab:example concepts,tab:example concepts IN100,tab:example concepts Food,tab:example concepts Flower}.

\subsection{Additional Implementation Details} \label{sec:Additional imp_details}

\minisection{More Details of Concept Selection Module.}
Following ~\cite{LM4CV}, we implemented the concept selection module (CS) as a simple MLP. At the beginning of task $t$, the CS is trained on the training data of current task with cross-entropy loss and Mahalanobis loss, encouraging CS to construct embedding space with vision-language knowledge. We train CS for 30 epochs with a batch size of 64 and a learning rate of 0.01 . After training, the learned weight matrix of CS is leveraged to select concepts from concepts set $\mathcal{C}_t$ of task $t$, based on the distances between text features $f_T(\mathcal{C}_t)$ and weight matrix.

\minisection{Pseudo Code.}
The training pipeline of our proposed method is outlined in \Cref{alg:algo}, the image encoder $f_I$ and text encoder $f_T$ of CLIP are both frozen during the whole training process, only the CBL, classifier and CS are trainable. For each task, we first select concepts to construct a bottleneck and extract prototypes for the categories relevant to the current task. Afterward, old prototypes are augmented using data from the current task to address catastrophic forgetting. Lastly, we calculate the CLIP concept activation matrix and train our model with the guidance of three loss functions as described in \Cref{mini:CA}.

\begin{algorithm}
\caption{\textsc{LG-CBM} for Interpretable CL}\label{alg:algo}
\textbf{Input:} Incremental datasets: $\{\mathcal{D}^1, \mathcal{D}^2, \cdots, \mathcal{D}^n\}$, Task-specific concepts: $\{\mathcal{C}_1, \mathcal{C}_2, \cdots, \mathcal{C}_n\}$, Pre-trained CLIP image encoder and text encoder: $f_I$, $f_T$. \\
\textbf{Output:} Incrementally trained model with interpretability
\begin{algorithmic}[1]
\For{$t = 1, 2, \cdots, n$}
    \State Get the training set $\mathcal{D}^t$ and Concepts $\mathcal{C}_t$;
    \State Extract text feature of $\mathcal{C}_t$;
    \State Select concepts from $\mathcal{C}_t$ to form bottlenecks $B_t$;\label{algo:bottleneck}
    \State Extract the prototypes of $\mathcal{D}^t$ as $P_t$;\label{algo:prototypes}
    \If{$t > 1$}
        \State Augment prototypes $P_{1:t-1}$ via \Cref{eq:semantic relation} 
        \Statex \hspace{3em}and \Cref{eq:augment prototypes};\label{algo: augmentation}
    \EndIf
    \State Compute CLIP concept activation matrix
    \Statex \hspace{1.5em}via \Cref{eq:clip_activation_matrix}; \label{algo:E_clip}
    \State Optimize the CBL and classifier via \Cref{eq:final loss}; \label{algo:loss}
\EndFor
\end{algorithmic}
\end{algorithm}

\subsection{More Results of Experiments} \label{sec: more exps}
\begin{figure*}[h]
    \centering
    \begin{subfigure}[t]{\textwidth}
        \centering
        \includegraphics[width=\textwidth]{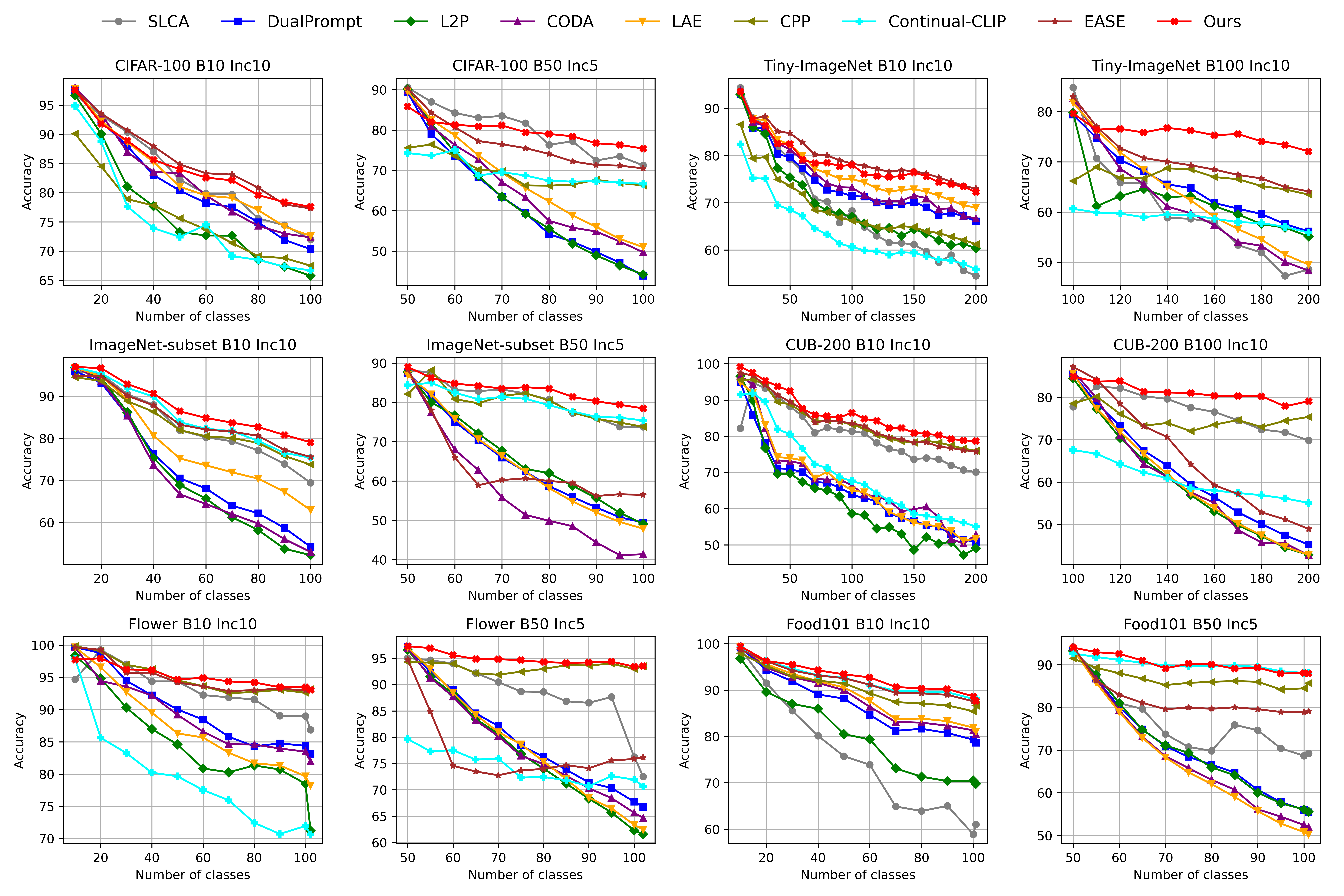}
    \end{subfigure}

    \begin{subfigure}[t]{\textwidth}
        \centering
        \includegraphics[width=0.5\textwidth]{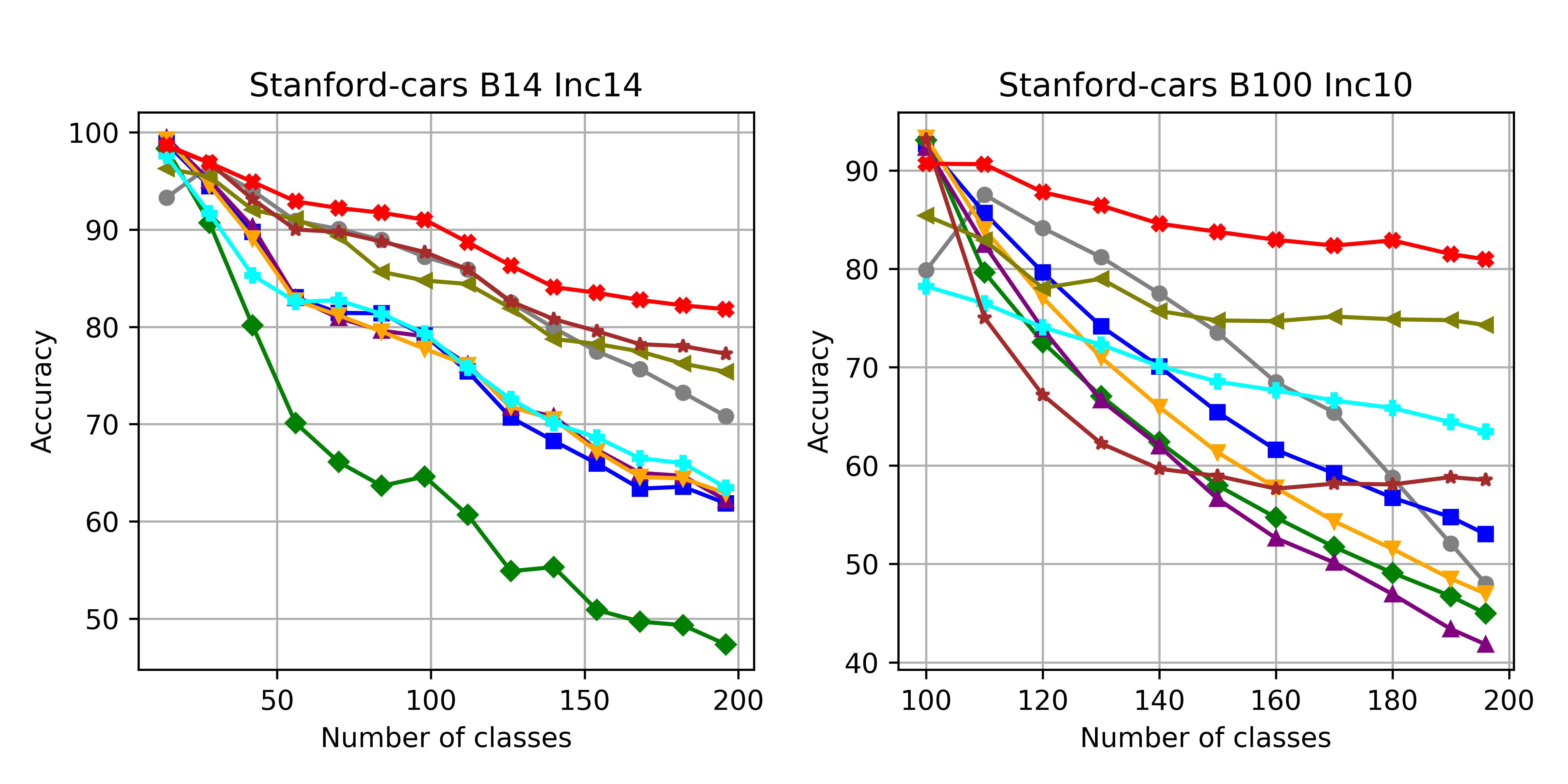}
    \end{subfigure}

    \caption{Average incremental accuracy $\bar{A}$ curve of our method with other state-of-the-art methods across all benchmark datasets.} 
    \label{fig:linesplot}
\end{figure*}

\minisection{Accuracy Curve in Various Settings.}
We illustrate the accuracy decreasing trends of our method with other state-of-the-art baselines on $\bar{A}$ across all benchmark datasets in \Cref{fig:linesplot}. 
Our method outperforms other methods with higher accuracy and less forgetting in most settings, especially when the initial task contains almost half of categories of the entire datasets, the $A_{last}$ performance of our method is also best on most benchmark datasets.

\minisection{Concepts From Different ChatGPT.}
As shown in \Cref{tab:gpt}, we report the average incremental accuracy $\bar{A}$ on ImageNet-subset and Flower datasets with concepts generated from different ChatGPT models. The findings demonstrate that concepts derived from distinct ChatGPT models have minimal impact on the model's performance, with accuracy fluctuations of less than 0.2\% on ImageNet-subset and 0.5\% on Flower. For all experiments conducted with our method, we obtain concepts by querying GPT-4o.

\begin{table}[h]
    \setlength{\abovecaptionskip}{2pt} 
    \setlength{\belowcaptionskip}{-3pt}
    \caption{The average incremental accuracy $\bar{A}$ on the ImageNet-subset and Flower datasets, evaluated using concepts derived from different versions of ChatGPT.}
    \centering
    \begin{tabular}{c c c}
    \toprule
         \multirow{2}{*}{\textbf{ChatGPTs}} 
         & \textbf{ImageNet-subset} & \textbf{Flower} \\
         & B10 Inc10 & B10 Inc10 \\
    \midrule
         GPT-3.5 turbo & 87.63 & 94.69  \\
         GPT-4 turbo & 87.50 & 94.76  \\
         GPT-4o & 87.55 & 95.16 \\
    \bottomrule
    \end{tabular}
    \label{tab:gpt}
\end{table}

\minisection{Performance in Non-continual Setting.}
We present the results for joint training of our method on three coarse-grained datasets in \Cref{tab: joint setting}. A performance gap remains between continual learning methods and joint training.

\begin{table}[h]
\centering
\caption{The performance of our method under Non-continual setting. }
\begin{adjustbox}{width=0.48\textwidth}
\begin{tabular}{c c c c}
\toprule

{\textbf{Methods}}  & {\textbf{CIFAR-100}} & {\textbf{Tiny-ImageNet}} & {\textbf{ImageNet-Subset}} \\

\midrule
\textbf{Joint training} & 82.14$\pm$0.02 & 76.43$\pm$0.11 & 83.52$\pm$0.06 \\
\bottomrule
\end{tabular}
\end{adjustbox}

\label{tab: joint setting} 
\end{table}

\minisection{Additional Ablation Study.}
Semantic-Guided Prototype Augmentation (PA) is designed to enhance knowledge retention. We evaluate its impact by comparing it against “Base” (without any anti-forgetting strategy) and “Base+Proto” (classical prototype loss without augmentation). The results show the effectiveness of the PA module.

\begin{table}[h]
\centering
\caption{The effectiveness of semantic-guide prototype augmentation module.}
\begin{adjustbox}{width=0.45\textwidth}
\begin{tabular}{c cc cc cc}
\toprule

 & \textbf{CIFAR-100} & \textbf{Tiny-ImageNet} & \textbf{ImageNet-Subset} \\
& B50 Inc5 & B100 Inc10 & B50 Inc5 \\

\midrule
Base &  11.02$\pm$1.42 & 12.96$\pm$3.97 & 23.90$\pm$8.50 \\ 
Base+Proto & 69.15$\pm$0.38 & 64.74$\pm$0.62 & 74.23$\pm$0.25 \\
Base+PA & \textbf{75.91$\pm$0.50} & \textbf{71.97$\pm$0.09} & \textbf{78.21$\pm$0.29} \\
\bottomrule
\end{tabular}
\end{adjustbox}

\label{tab:PA_ablation} 
\end{table}

\noindent The Concept Alignment (CA) module helps learn human-understandable bottlenecks and enhances interpretability. We show an example of ``Azalea'' to analyze CA, listing the top-5 concepts in \Cref{tab: CA_ablation}. With CA, the selected concepts are all positive and align well with human understanding, emphasizing the importance of the CA module.

\begin{table}[ht]
\centering
\caption{The comparison of Top-5 concepts that contribute most to classify ``Azalea'' with and without CA module.}
\begin{adjustbox}{width=0.48\textwidth}
\begin{tabular}{c c c}
\toprule
{\textbf{Image}} & {\textbf{CA}} & {\textbf{Top-5 Concepts}}  \\

\midrule
\multirow{10}{*}{\includegraphics[width=0.15\linewidth]{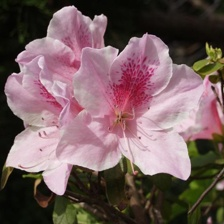}} & \multirow{5}{*}{w/o CA} 
& \textbf{NOT} radiate from the center like rays of sunshine \\ 
& & each petal is thin and almost translucent \\
& & \textbf{NOT} encase a deep pink center \\
& & \textbf{NOT} striking symmetry radiating from the center \\
& & stark contrast against its dark green foliage \\
\cmidrule(lr){2-3}
& \multirow{5}{*}{w/ CA}
& support small pink blossoms \\ 
& & bright pink in color \\
& & vibrant pink hue with a pale white margin \\
& & thrive in the shaded nooks of the tree limbs \\
& & stark contrast against its dark green foliage \\

\bottomrule
\end{tabular}
\end{adjustbox}

\label{tab: CA_ablation} 
\end{table}

\subsection{Computational Overhead of Concept Selection Module} \label{sec: CS cost}
The computational overhead of the CS on three coarse-grained datasets is shown in \Cref{tab:training cost of CS}. We can infer from \Cref{tab:training cost of CS} that the training of CS is quite efficient, with low memory usage and computation, and the training time is also acceptable.

\begin{table}[h] 
\centering
\caption{The training cost of the concept selection module.}
\begin{adjustbox}{width=0.48\textwidth}
\begin{tabular}{c c c c}
\toprule
{\textbf{Metrics}}  & \textbf{CIFAR-100} & \textbf{Tiny-ImageNet} & \textbf{ImageNet-Subset} \\

\midrule
FLOPs (M) & 0.68 & 1.56 & 0.99 \\
Time per epoch (s) & 1.30 & 1.26 & 2.24 \\
Peak Memory Usage (MB) & 2571 & 2701 & 2673 \\
\bottomrule
\end{tabular}
\end{adjustbox}
\label{tab:training cost of CS} 
\vspace{-3mm}
\end{table}

\subsection{More Interpretable Model Predictions.} \label{sec: more interpretable preds}
\begin{figure*}[h]
    \centering
    \begin{subfigure}[t]{\textwidth}
        \centering
        \includegraphics[width=\textwidth]{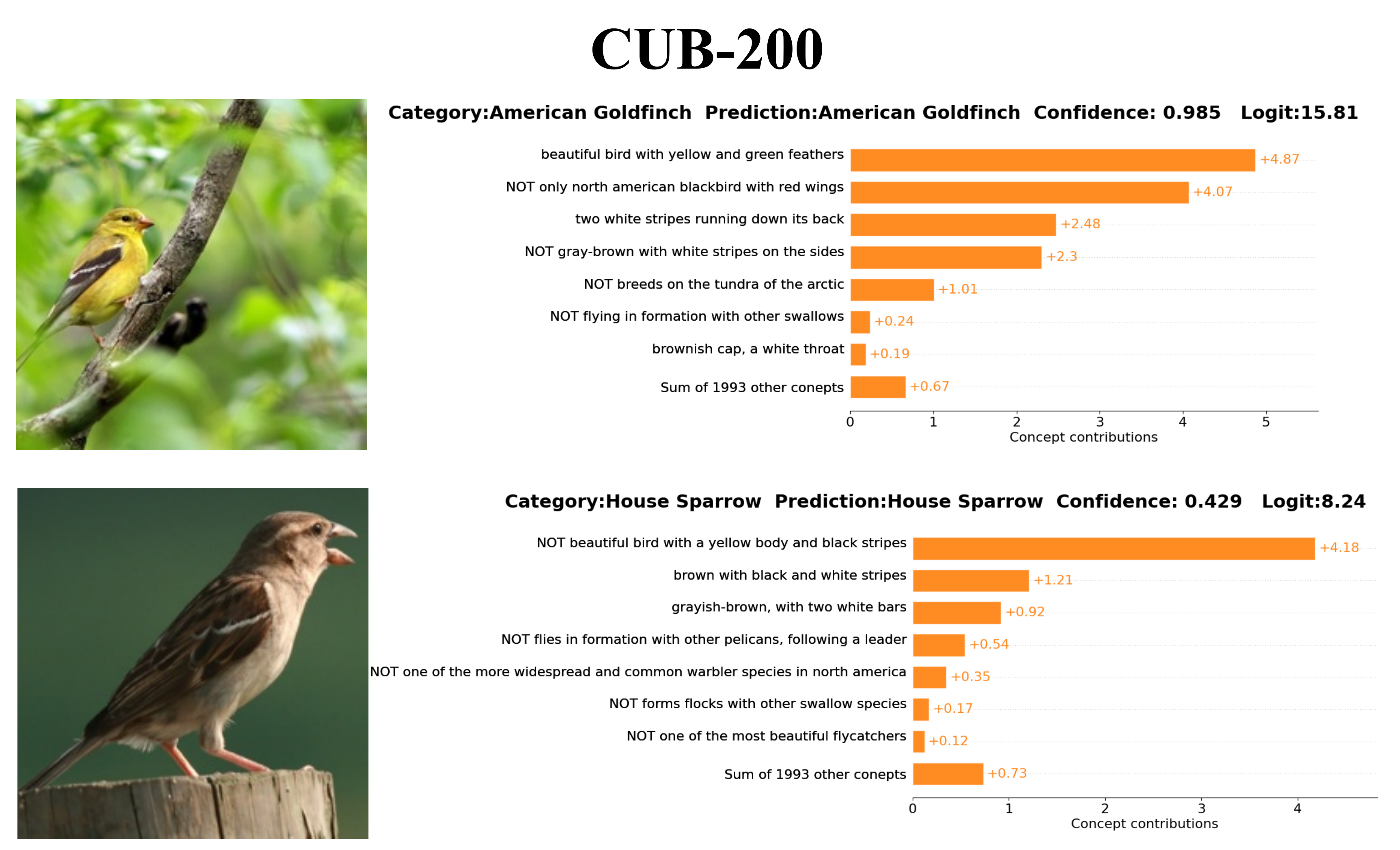}
    \end{subfigure}

    \begin{subfigure}[t]{\textwidth}
        \centering
        \includegraphics[width=\textwidth]{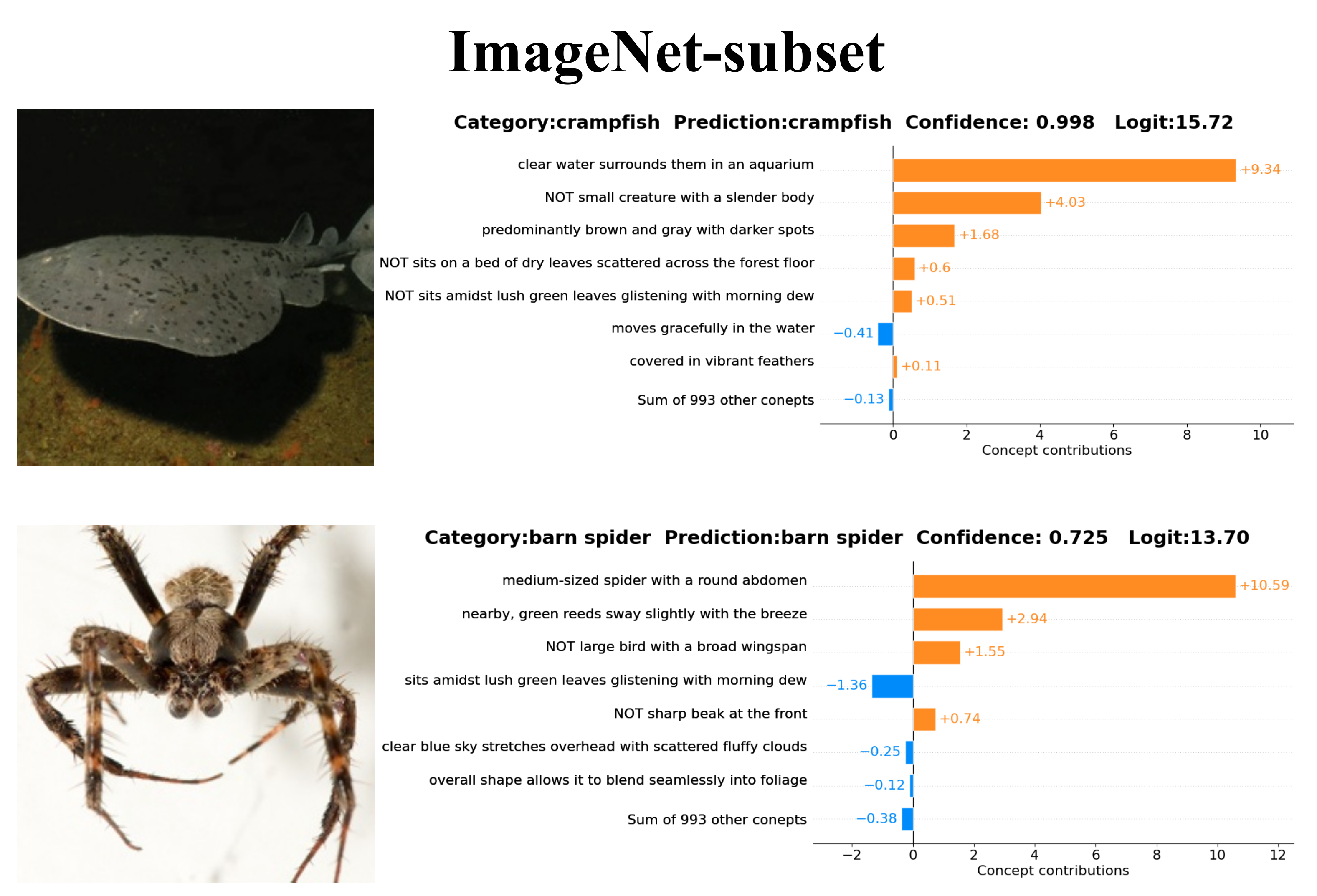}
    \end{subfigure}
    \caption{Contribution Visualization after training on CUB-200 B10 Inc10 and ImageNet-Subset B10 Inc10.}
    \label{fig:CV-CUB&IN100-more}
\vspace{-4mm}
\end{figure*}

\begin{figure*}[h]
    \centering
    \begin{subfigure}[t]{\textwidth}
        \centering
        \includegraphics[width=\textwidth]{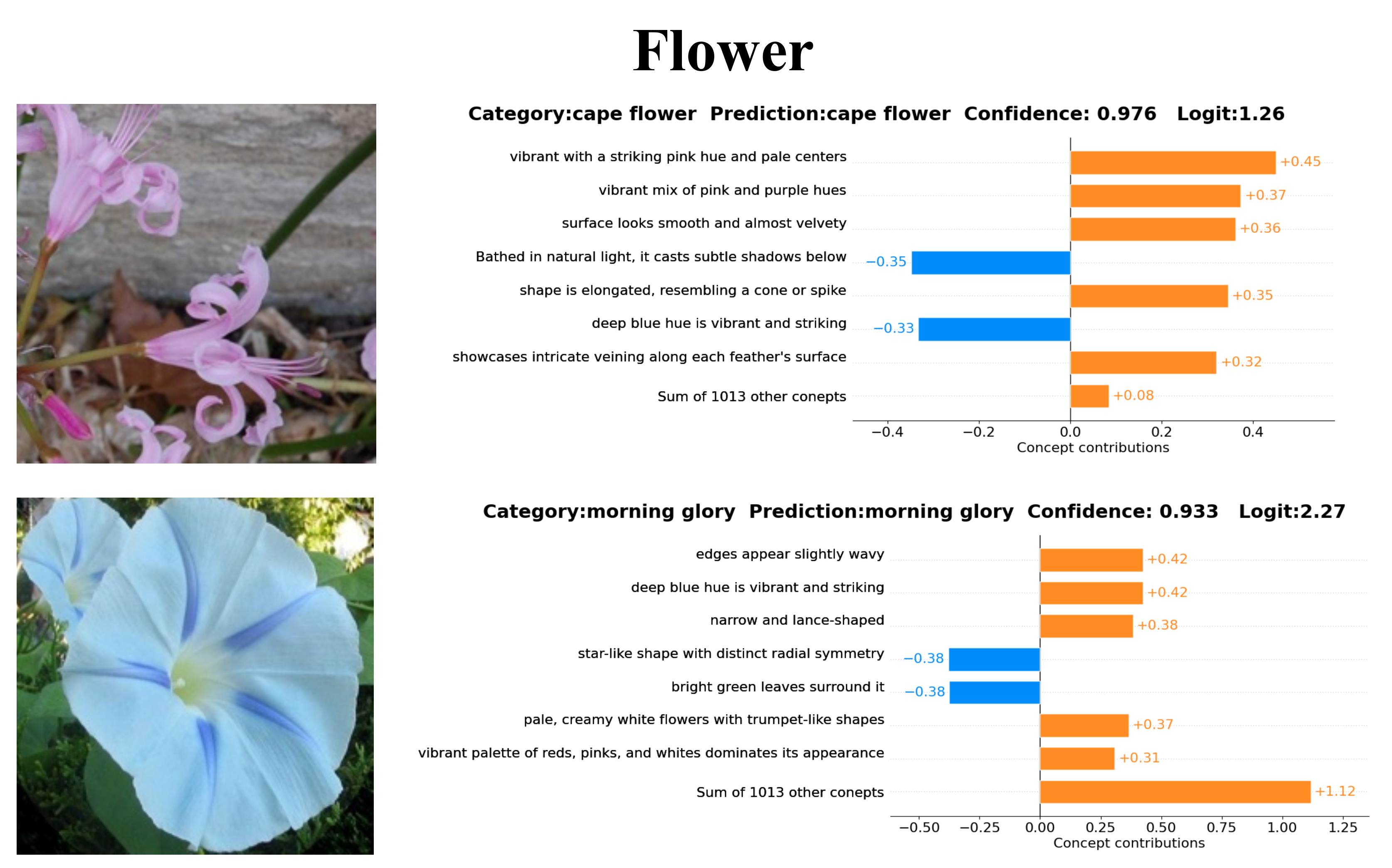}
    \end{subfigure}
    \begin{subfigure}[t]{\textwidth}
        \centering
        \includegraphics[width=\textwidth]{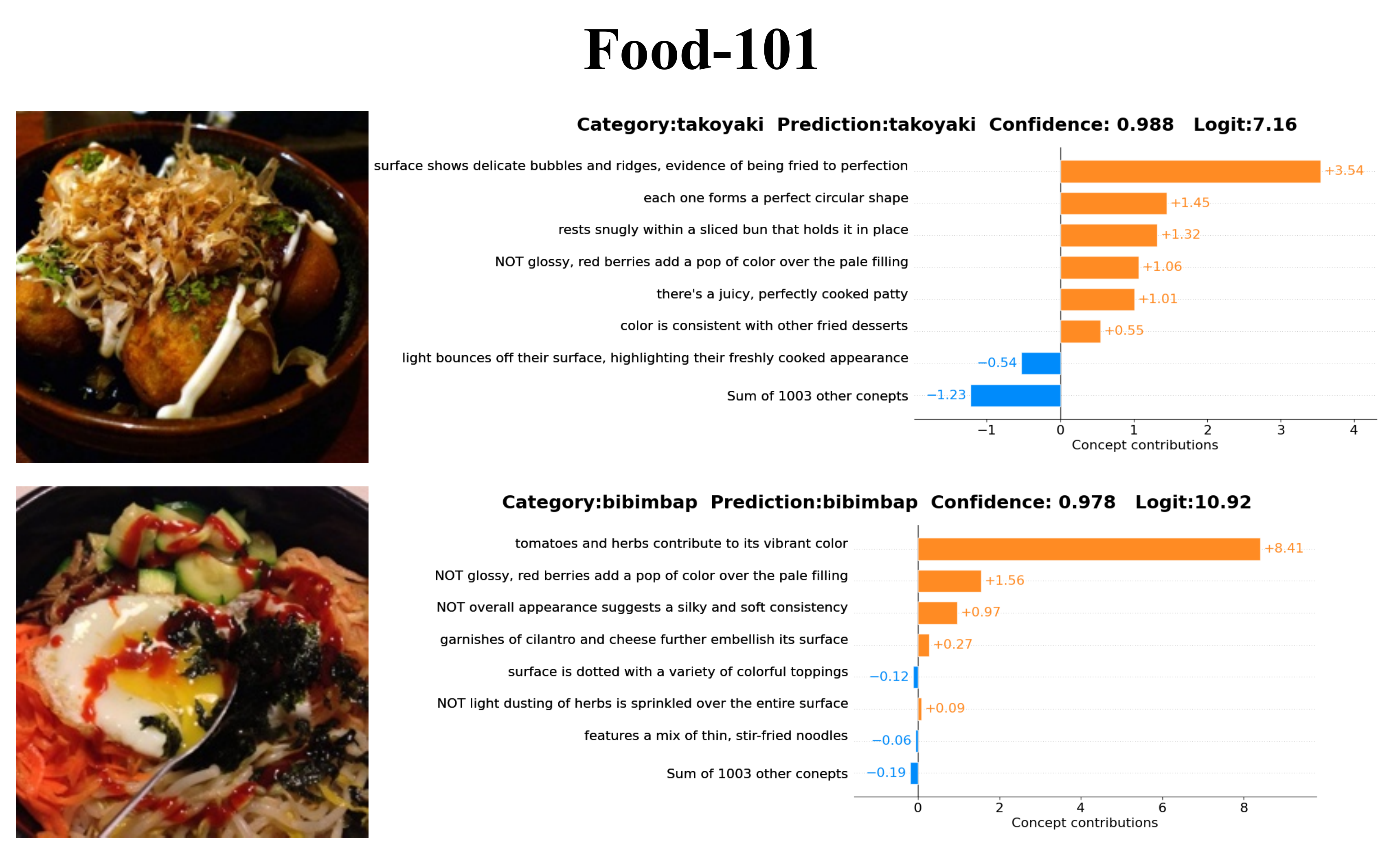}
    \end{subfigure}
    \caption{Contribution Visualization after training on Flower B10 Inc10 and Food-101 B10 Inc10.}
    \label{fig:CV-FLower&Food-more}
\vspace{-4mm}
\end{figure*}

\begin{figure*}[h]
    \setlength{\abovecaptionskip}{10pt} 
    \setlength{\belowcaptionskip}{-10pt}
    \centering
    \includegraphics[width=\textwidth]{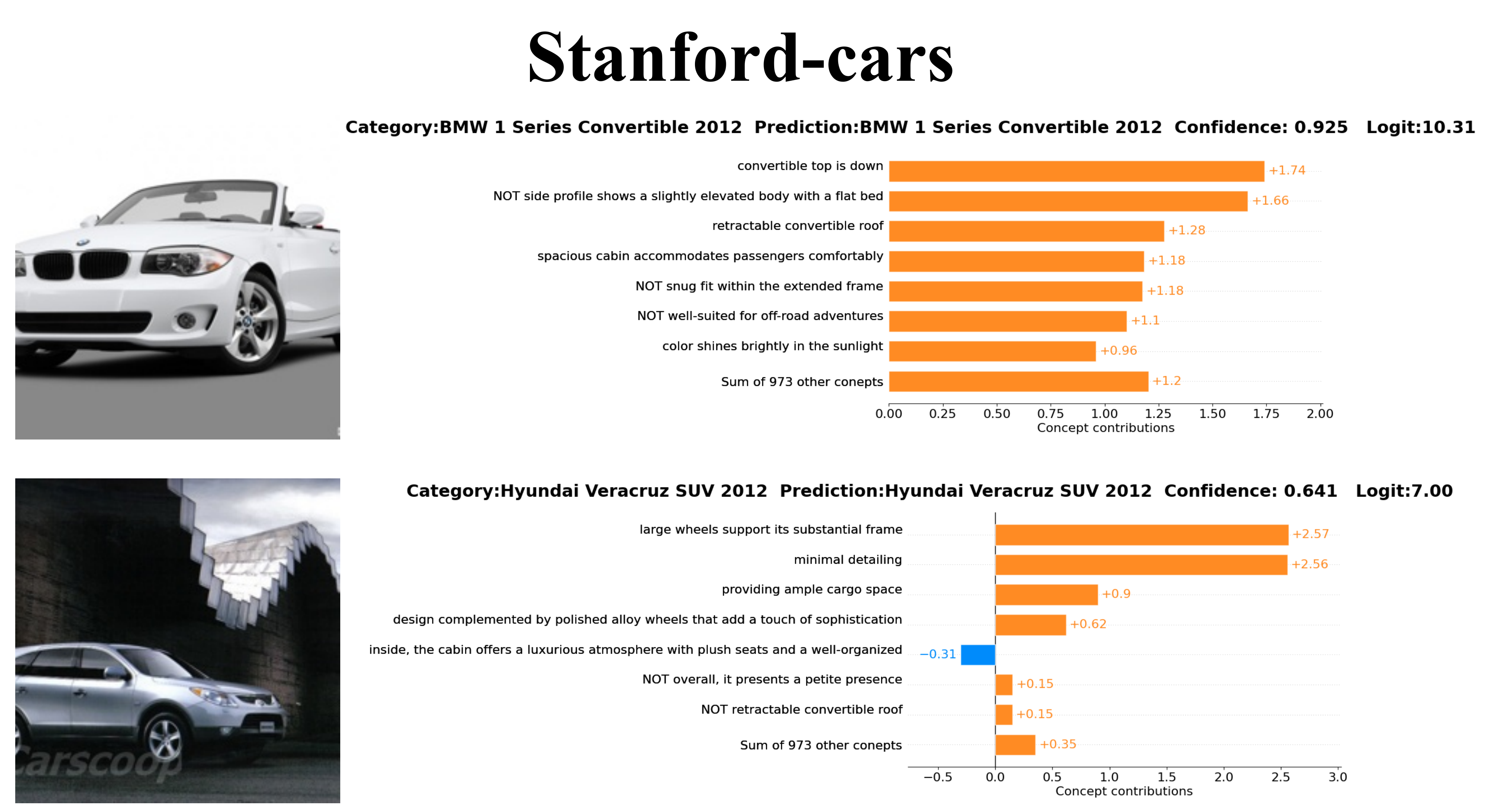}
    \caption{Contribution Visualization after training on Stanford-cars B14 Inc14.}
    \label{fig:CV-cars-more}
\vspace{-4mm}
\end{figure*}

As depicted in \Cref{fig:CV-CUB&IN100-more,fig:CV-FLower&Food-more,fig:CV-cars-more}, we provide additional examples of interpretable model predictions across various datasets. We can find that our method demonstrates excellent interpretability across all benchmark datasets, delivering strong performance accompanied by coherent and logical explanations.

\begin{table*}[h]
\centering
\captionsetup[table]{justification=centering, singlelinecheck=false} 
\large
\caption{Example concepts of categories in CUB-200.}
\begin{adjustbox}{width=\textwidth}
\begin{tabular}{c l c l}
\toprule
\textbf{Category} & \textbf{Concepts} & \textbf{Category} & \textbf{Concepts} \\
\midrule
Black footed Albatross
    & \makecell[l]{``graceful, elongated body", \\
     ``long, slender wings",\\
     ``plume is primarily dark brown with subtle shades of gray",\\
     ``pale patch on its face creates an interesting visual effect",\\
     ``distinguishing feature against the rest of its body",\\
     ``large wingspan that stretches wide",\\
     ``body appears robust and streamlined",\\
     ``strong and hooked",\\
     ``overall form is well-suited for gliding over the ocean",\\
     ``endless, clear sky",\\
     ``glides over the gentle waves with ease"\\}
& Crested Auklet
    & \makecell[l]{``dark, slate-gray body",\\
     ``sleek look",\\
     ``head is adorned with a striking crest",\\
     ``adds to its distinctive appearance",\\
     ``around its beak",\\
     ``there is a splash of vibrant orange",\\
     ``small, bright eyes stand out against the darker feathers",\\
     ``small bird with a compact body",\\
     ``beak is short and slightly curved",\\
     ``prominent crest on its head",\\
     ``relatively short compared to its body",\\
     ``rocky coastline is covered in patches of green moss",\\
     ``waves crash against the shore under an overcast sky"}\\
\midrule
Rusty Blackbird
    & \makecell[l]{``dark, glossy plumage that shimmers in the sunlight",\\
     ``feathers display a subtle iridescence of blues and greens",\\
     ``rusty hue adorns its wings and patches around its eyes",\\
     ``bill is slender and pointed",\\
     ``suitable for foraging",\\
     ``medium-sized bird with a slender build",\\
     ``body appears elongated and streamlined",\\
     ``has a relatively long tail",\\
     ``straight and slightly pointed",\\
     ``bird is perched on a thin branch, surrounded by dense foliage",\\
     ``background features a calm stream reflecting the overhanging trees"}

& Gray Catbird
    & \makecell[l]{``sleek body with a predominantly gray plumage",\\
     ``head is capped with a darker, almost black hue",\\
     ``small, slender beak complements its smooth feathers",\\
     ``tail and wings show subtle, lighter shading",\\
     ``medium-sized bird with a slender body",\\
     ``overall shape is sleek and elongated",\\
     ``long tail that tapers towards the end",\\
     ``bird features a rounded head with a short, straight bill",\\
     ``gentle stream runs nearby",\\
     ``surrounded by tall reeds and vibrant autumn leaves"\\}\\
\bottomrule
\end{tabular}
\end{adjustbox}

\label{tab:example concepts} 
\end{table*}

\begin{table*}[h]
\centering
\captionsetup[table]{justification=centering, singlelinecheck=false} 
\caption{Example concepts of categories in ImageNet-subset.}
\begin{adjustbox}{width=\textwidth}
\begin{tabular}{c l c l}
\toprule
\textbf{Category} & \textbf{Concepts} & \textbf{Category} & \textbf{Concepts}  \\
\midrule
Goldfish
        & \makecell[l]{``bright orange coloration with a metallic sheen",\\
         ``fins are long and delicate",\\
         ``flowing gracefully in the water",\\
         ``sleek and oval-shaped",\\
         ``slight bulge at the sides",\\
         ``eyes are round and prominent",\\
         ``have an elongated body shape",\\
         ``typically around four to six inches long",\\
         ``fins are delicate and fan-like",\\
         ``bodies are often plump and streamlined",\\
         ``clear water surrounds them in an aquarium",\\
         ``green aquatic plants sway gently nearby"}
& Cock
        & \makecell[l]{``vibrant plumage with a mix of rich, earthy tones",\\
        ``bright red adorns its comb and wattle, adding a striking contrast",\\
        ``hues of green and blue",\\
        ``sturdy beak and strong legs complete the appearance",\\
        ``appears quite large in the photo",\\
        ``structure is straight and firm",\\
        ``surface looks smooth and uniform",\\
        ``overall, it gives an impression of solidity and symmetry",\\
        ``warm glow over the farmyard",\\
        ``illuminating the rustic wooden fence and scattered hay",\\
        ``nearby, a few chickens peck at the ground"\\}\\
\midrule
Tailed Frog
        & \makecell[l]{``brown, mottled skin",\\
        ``blends with the forest floor",\\
        ``large and sit prominently on its head",\\
        ``muscular and well-developed for jumping",\\
        ``small, tail-like appendage extends from its rear",\\
        ``small and robust with a rounded body",\\
        ``legs appear muscular and strong",\\
        ``head is wide with prominent eyes",\\
        ``slender tail extends from its back",\\
        ``small amphibian sits among the damp, moss-covered rocks",\\
        ``lush, green environment"}
& Agama
        & \makecell[l]{``vibrant body with a mixture of colors",\\
        ``head is often bright red or orange",\\
        ``body can display hues of blue or brown",\\
        ``smooth and glossy under the light",\\
        ``medium-sized reptile with a stout body",\\
        ``limbs are strong and muscular, aiding in movement",\\
        ``head is somewhat triangular, tapering towards snout",\\
        ``tail is long and slender, extending beyond the body",\\
        ``rocky terrain is dotted with patches of dry grass and scattered stones",\\
        ``clear blue sky stretches above",\\
        ``casting shadows on sunlit ground"\\}\\
\bottomrule
\vspace{-5mm}
\end{tabular}
\end{adjustbox}
\label{tab:example concepts IN100} 
\end{table*}

\begin{table*}[h]
\large
\centering
\captionsetup[table]{justification=centering, singlelinecheck=false} 
\caption{Example concepts of categories in Food101.}
\label{tab:example concepts Food} 
\begin{adjustbox}{width=\textwidth}
\begin{tabular}{c l c l}
\toprule
\textbf{Category} & \textbf{Concepts} & \textbf{Category} & \textbf{Concepts}  \\
\midrule
Chocolate Cake
        &\makecell[l]{``appears round and sits on a flat surface",\\
        ``rich, dark brown",\\
        ``smooth, glossy finish",\\
        ``reflect off its slightly uneven edges",\\
        ``uniform throughout",\\
        ``appears dense yet tender",\\
        ``surface shows a slight sheen, suggesting a moist consistency",\\
        ``each slice reveals layers with a slightly crumbly edge;",\\
        ``smoothly glazed",\\
        ``contrasting with the softer interior",\\
        ``Tiny air pockets are scattered throughout",\\
        ``spongy nature",\\
        ``reveals a dense and moist interior",\\
        ``texture appears soft, showcasing fine crumbs",\\
        ``uniform color throughout",\\
        ``Tiny flecks may indicate the presence of fine ingredients"}
& Hot dog
        &\makecell[l]{``long, cylindrical shape with slightly rounded ends",\\
        ``nestled within a soft, oblong bun",\\
        ``bright red and mustard streaks on top",\\
        ``suggest a freshly cooked state",\\
        ``cooked evenly, its texture looks firm yet somewhat flexible",\\
        ``glisten, indicating juiciness",\\
        ``densely packed interior",\\
        ``hinting at a substantial bite",\\
        ``inside, it reveals a grilled sausage nestled within the bun",\\
        ``slightly browned",\\
        ``shows a textured casing",\\
        ``drizzled atop",\\
        ``blending into the sausage",\\
        ``soft layered",\\
        ``provide a cozy embrace"\\}\\
\midrule
Lasagna
        &\makecell[l]{``rectangular dish layered with alternating levels of ingredients",\\
        ``golden-brown with a slightly crisp texture",\\
        ``edges appear slightly darker, indicating a well-cooked surface",\\
        ``visible stripes of red and white sauce peeking through through",\\
        ``warm and inviting look",\\
        ``various shades of red, brown, and cream",\\
        ``layered structure with firm and slightly chewy pasta sheets",\\
        ``gooey and stretchy consistency from the melted cheese",\\
        ``moist and saucy texture between the layers",\\
        ``blend of soft, meaty, and creamy elements within it",\\
        ``visible layers alternate between creamy white and tangy red sauces",\\
        ``oozes over the top, creating a slightly golden crust",\\
        ``add texture and flavor",\\
        ``thin, flat pasta shapes the structure, holding everything together",\\
        ``specks of green are scattered throughout"\\}
& Ice-cream
        &\makecell[l]{``smooth, round shape with slight indentations",\\
        ``hue blends subtle shades of beige and light brown",\\
        ``surface appears glossy, catching light reflections",\\
        ``small, dark specks adds texture detail to its appearance",\\
        ``contrasts sharply with the dark background",\\
        ``smooth, creamy surface",\\
        ``glistens under the light",\\
        ``heaped scoops reveal tiny air pockets throughout",\\
        ``small droplets of condensation are visible on its exterior",\\
        ``soft, slightly elastic quality when scooped",\\
        ``melting edges give way to a glossy, liquid sheen",\\
        ``smooth and rich",\\
        ``Tiny air bubbles are evenly dispersed throughout the frozen treat",\\
        ``small vanilla bean specks are visible, hinting at quality ingredients",\\
        ``surface has a slightly glistening, frosty appearance",\\
        ``dense, yet soft composition"\\}\\
\bottomrule
\end{tabular}
\end{adjustbox}
\end{table*}

\begin{table*}[h]
\centering
\captionsetup[table]{justification=centering, singlelinecheck=false} 
\caption{Example concepts of categories in Flower.}
\label{tab:example concepts Flower} 
\begin{adjustbox}{width=\textwidth}
\begin{tabular}{c l c l}
\toprule
\textbf{Category} & \textbf{Concepts} & \textbf{Category} & \textbf{Concepts}  \\
\midrule
Fire lily
        &\makecell[l]{``vibrant red and orange petals that flare outward",\\
        ``petals have a wavy and slightly ruffled texture",\\
        ``long, green stamens protrude prominently at the center",\\
        ``leaves are slender and gracefully arch away from the stem",\\
        ``sits atop a tall, curved stem",\\
        ``adds elegance to its appearance",\\
        ``vivid shade of red-orange",\\
        ``each petal gracefully curls backwards",\\
        ``bright yellow",\\
        ``long, slender stamens protrude from it",\\
        ``overall shape is elegant and delicate",\\
        ``vibrant orange-red petals that curve gracefully backwards",\\
        ``stamens and pistil extend prominently from the center",\\}
& Red ginger
        &\makecell[l]{``stands tall with elongated, vibrant red bracts crowded closely together",\\
        ``strong green stem supports the structure and extends upwards",\\
        ``leaves are broad, glossy, and waxy with a deep green color",\\
        ``graceful arcs",\\
        ``add to its elegance",\\
        ``striking combination of vivid color and slender form",\\
        ``vibrant red color",\\
        ``stands out vividly",\\
        ``shape is elongated, resembling a cone or spike",\\
        ``smooth",\\
        ``layered neatly around the core",\\
        ``green leaves surround its base, framing it beautifully",\\}\\
\midrule
Corn poppy
        &\makecell[l]{``vibrant red petals that catch the eye",\\
        ``dark and contrasting, almost black",\\
        ``stem is slender and green, standing tall",\\
        ``sparse and slightly jagged",\\
        ``overall appearance is delicate yet striking",\\
        ``vibrant with a bright red hue",\\
        ``petals are delicate and slightly crinkled",\\
        ``flower has a central dark spot marking its contrast",\\
        ``stands tall on a slender green stem",\\
        ``overall shape is rounded and open",\\
        ``petals are bright red and slightly crinkled",\\
        ``black spot marks the base of each petal",\\}
& Artichoke
        &\makecell[l]{``stands tall with thick, green stems reaching upwards",\\
        ``leaves are large, broad",\\
        ``have a silvery hue",\\
        ``flower head is a round, dense cluster of tightly layered bracts",\\
        ``hint of purple peek through as the flower begins to bloom",\\
        ``delicate thorns edge the tips of its protective leaves",\\
        ``vibrant purple hue",\\
        ``stands tall with a robust structure",\\
        ``petals form spherical shape",\\
        ``petals overlap tightly together",\\
        ``overall appearance is spiky yet symmetrical",\\
        ``large, spiky petals",\\
        ``curl out in a vibrant display",\\}\\
\bottomrule
\vspace{-5mm}
\end{tabular}
\end{adjustbox}
\end{table*}

\end{document}